\RequirePackage{fix-cm}

\documentclass[twoside,11pt]{article}
\usepackage{jair, theapa, rawfonts}

\usepackage{amsmath}
\usepackage{amssymb}
\usepackage{bm}
\usepackage{graphicx}
\usepackage{epstopdf}
\usepackage{algorithm}
\usepackage[noend]{algpseudocode}
\usepackage{multirow}
\usepackage{pgfplots}
\usepackage{pgfplotstable}
\usepackage{tikz}
\usetikzlibrary{arrows.meta}
\usepgfplotslibrary{groupplots}
\usepackage{lscape}
\usepackage{siunitx}
\usepackage{booktabs}
\usepackage[font=small,skip=5pt]{caption}
\usepackage[hyphens]{url}
\usepackage{subcaption}
\usepackage{tabularx}
\usepackage{boondox-cal} 

\DeclareCaptionLabelFormat{andtable}{#1~#2  \&  \tablename~\thetable}

\usetikzlibrary{bayesnet}

\newcommand{\citep}[1]{\cite{#1}}
\newcommand{\citet}[1]{\citeauthor{#1} \citeyear{#1}}

\pgfplotsset{compat=1.11}
\pgfplotsset{
    jitter/.style={
        x filter/.code={\pgfmathparse{\pgfmathresult+rnd*#1}}
    },
    jitter/.default=0.1
}

\newcommand{\bigo}{O} 

\newcommand{\paren}[1]{\mathopen{}\mathclose\bgroup\left(#1\aftergroup\egroup\right)}
\newcommand{\brock}[1]{\mathopen{}\mathclose\bgroup\left[#1\aftergroup\egroup\right]}
\newcommand{\curly}[1]{\mathopen{}\mathclose\bgroup\left\{#1\aftergroup\egroup\right\}}
\newcommand{\anglebrackets}[1]{\langle #1 \rangle}

\DeclareMathOperator{\Pa}{pa} 
\DeclareMathOperator{\Ch}{ch} 
\newcommand{\cX}{\mathcal{X}} 
\newcommand{\cx}{\mathcal{x}} 
\newcommand{\cu}{\mathcal{u}} 
\newcommand{\maxpasses}{\hat{n}_\text{cycle}}
\newcommand{\discset}{\Lambda_{\bm{\cX}}}

\jairheading{59}{2017}{103-132}{10/16}{06/17}
\ShortHeadings{Learning Discrete Bayesian Networks from Continuous Data}
{Chen, Wheeler, \& Kochenderfer}
\firstpageno{103}

\begin{document}

\title{Learning Discrete Bayesian Networks from Continuous Data
}


\author{\name Yi-Chun Chen \email yichunc0419@alumni.stanford.edu \\
	\addr Institute of Computational and Mathematical Engineering,\\
	Stanford University, Stanford, CA  94035 USA
	\AND
	\name Tim A. Wheeler \email wheelert@stanford.edu \\
	\addr Department of Aeronautics and Astronautics,\\ Stanford University,
	Stanford, CA  94035 USA
	\AND
	\name Mykel J. Kochenderfer \email mykel@stanford.edu \\
	\addr Department of Aeronautics and Astronautics,\\ Stanford University,
	Stanford, CA  94035 USA}

\maketitle

\begin{abstract}

Learning Bayesian networks from raw data can help provide insights into the relationships between variables. 
While real data often contains a mixture of discrete and continuous-valued variables, many Bayesian network structure learning algorithms assume all random variables are discrete.
Thus, continuous variables are often discretized when learning a Bayesian network.
However, the choice of discretization policy has significant impact on the accuracy, speed, and interpretability of the resulting models.
This paper introduces a principled Bayesian discretization method for continuous variables in Bayesian networks with quadratic complexity instead of the cubic complexity of other standard techniques.
Empirical demonstrations show that the proposed method is superior to the established minimum description length algorithm.
In addition, this paper shows how to incorporate existing methods into the structure learning process to discretize all continuous variables and simultaneously learn Bayesian network structures.

\end{abstract}


\section{Introduction}

Bayesian networks \citep{Pearl_1988,PGM_2009} are often used to model uncertainty and causality, with applications ranging from decision-making systems \citep{kochenderfer2012next} to medical diagnosis \citep{Lustgarten_2011}. Learning Bayesian networks from raw data can help users discover knowledge without field-specific experts, such as causality and probabilistic relation between variables. It is common to assume that the random variables in a Bayesian network are discrete, since many Bayesian network learning algorithms are unable to efficiently handle continuous variables. In addition, many of the commonly used Bayesian network software packages, such as Netica \citep{netica1992}, SMILearn \citep{druzdzel1999smile}, and bnlearn \citep{bnlearn2010}, are designed for discrete variables. However, many applications require the use of continuous data, such as positions and velocities in dynamic systems \citep{kochenderfer2010airspace}.

There are three common approaches to extending Bayesian network learning algorithms to continuous variables. The first is to model the conditional probability density of each continuous variable using specific families of parametric distributions, and then to redesign Bayesian network learning algorithms based on the parameterizations. One example of parametric continuous distributions in Bayesian networks is the Gaussian graphical model \citep{Weiss_2011}.

The second approach is to use nonparametric distributions, such as particle representations and Gaussian processes \citep{Ickstadt_2010}.
Unlike parametric methods, nonparametric methods can often fit any underlying probability distribution given sufficient data.
Parametric models have a fixed number of parameters, whereas the number of parameters in a nonparametric model grow with the amount of training data.

The third approach is discretization.
Automated discretization methods have been studied in machine learning and statistics for many years \citep{Dougherty_1995,Kerber_1992,Holte_1993,Fayyad_1993}, primarily for classification problems.
These methods learn the best discretization policy for a continuous attribute by considering its interaction with a class variable.
It is common to discretize all continuous variables before learning a Bayesian network structure to avoid having to consider variable interactions, as the interactions and dependencies between variables in Bayesian networks introduce complexity.
Prior work exists for discretizing continuous variables in naive Bayesian networks and tree-augmented networks \citep{Fried_naive}, but only a few discretization methods for general Bayesian networks have been proposed \citep{Friedman_1996,Kozlov_1997,Monti_1998,Steck_2007}.

A common discretization method for Bayesian networks is to split continuous data into uniform-width intervals or to use field-specific expertise.
Several Bayesian methods exist to discretize according to the marginal distribution \citep{scott1979optimal,freedman1981histogram,knuth2013optimal,scargle2013studies}, but this ignores dependencies between variables and leads to suboptimal performance, as demonstrated in Appendix A.
A more principled method is the minimum description length (MDL) principle discretization \citep{Friedman_1996}.
The MDL principle proposed by \citet{MDL_1978} states that the best model for a dataset is the one that minimizes the amount of information needed to describe it \citep{Grunwald_2009}.
MDL methods trade off goodness-of-fit against model complexity to reduce generalization error.
In the context of Bayesian networks, \citet{Friedman_1996} applied the MDL principle to determine the optimal number of discretization intervals for continuous variables and the optimal positions of their discretization edges.
Their approach selects a discretization policy that minimizes the sum of the description length of the discretized Bayesian network and the information necessary for recovering the continuous values from the discretized data.

The optimal discretization policy of a single continuous variable under MDL can be found using dynamic programming in cubic runtime with the number of data instances.
For Bayesian networks with multiple continuous variables, MDL discretization can be iteratively applied to each continuous variable.
Only one variable is treated as continuous at a time, while all other continuous variables are treated as discretized based on an initial discretization policy or the discretization result from a previous iteration.

MDL discretization requires that the network structure be known in advance, but an iterative approach allows it to be incorporated into the structure learning process.
Simultaneous structure learning and discretization alternates between traditional discrete structure learning and optimal discretization.
Starting with some preliminary discretization policy, one first applies a structure learning algorithm to identify the locally optimal graph structure.
One then refines the discretization policy based on the learned network.
The cycle is repeated until convergence.

Results in this work suggest that the MDL method suffers from low sensitivity to discretization edge locations and returns too few discretization intervals for continuous variables.
This is caused by MDL's use of mutual information to measure the quality of discretization edges.
Mutual information, which is composed of empirical probabilities computed using event count ratios, varies less significantly with the positions of discretization edges than the method we suggest in this article.

MODL \citep{Boulle_2006} is a Bayesian method for discretizing a continuous feature according to a class variable, which selects the model with maximum probability given the data.
The MODL method uses dynamic programming to find the optimal discretization policy for a continuous variable given a discrete class variable, and has an $\bigo\paren{n^3 + r \cdot n^2}$ runtime, where $r$ is the number of class variable instantiations.
\citet{Lustgarten_2011} suggested several formulations for the prior over models. The asymptotic equivalence between MDL and MODL on the single-variable, single-class problem was examined by \citet{VL_2000}.

This paper describes a new Bayesian discretization method for continuous variables in Bayesian networks, extending prior work on single-variable discretization methods from \citet{Boulle_2006} and \citet{Lustgarten_2011}.
The proposed method optimizes the discretization policy relative to the network and takes parents, children, and spouse variables into account.
The optimal single-variable discretization method is derived in Section~\ref{sec:single_var}, using a prior which reduces the discretization runtime to $\bigo\paren{r \cdot n^2}$ without sacrificing optimality.
Section~\ref{sec:multi_var} covers Bayesian networks with multiple continuous variables and Section~\ref{sec:structure_learning} covers discretization while simultaneously learning network structure.
The paper concludes with a comparison against the existing minimum-description length \citep{Friedman_1996} method on real-world and synthetic datasets in Section~\ref{sec:experiments}.


\section{Preliminaries}
\label{sec:preliminaries}
This section covers the notation used throughout the paper to describe discretization policies.
This section also provides a brief overview of Bayesian networks.

\subsection{Discretization Policies}
\label{subsec:disc_policy}

Let $\cX$ be a continuous variable and let $\cx$ be a specific instance of $\cX$.
A discretization policy $\Lambda_{\cX} = \anglebrackets{e_1 < e_2 < \ldots < e_{k-1}}$ for $\cX$ is a mapping from $\mathbb{R}$ to $\curly{1,2,\ldots,k}$ such that
\begin{equation}
  \Lambda_{\cX}(\cx) = \begin{cases}
    1, & \text{if $x<e_1$}\\
    i, & \text{if $e_{i-1} \leq x < e_i$}\\
    k, & \text{otherwise.}
  \end{cases}
\end{equation}
The discretization policy discretizes $\cX$ into $k$ intervals.
Let the samples of $\cX$ in a given dataset $D$ be sorted in ascending order, ${\cx_{1:n} = \curly{\cx_1 \leq \cx_2 \leq \ldots \leq \cx_n}}$, and let the unique values be ${\cu_{1:m} = \curly{\cu_1 < \cu_2 < \ldots < \cu_{m}}}$.
The index of the last occurrence of $\cu_i$ in $\cx_{1:n}$ is denoted $s_i$.

The discretization edges $e_{1:k-1}$ mark the boundaries between discretization intervals.
In this paper, as with MODL, they are restricted to the midpoints between unique ascending instances of $\cX$.
Thus, each edge $e_i$ equals $\paren{\cu_{j} + \cu_{{j+1}}}/2$ for some $j$.
Two useful integer representations of $\Lambda_{\cX}$ can be written
\begin{equation}
\label{eq:disc_def}
  \Lambda_{\cX} = \anglebrackets{\lambda_1 < \lambda_2  < \ldots < \lambda_k} \equiv \anglebrackets{\gamma_1, \gamma_2, \ldots, \gamma_k}\text{,}
\end{equation}
where $\lambda_1 = s_1$, $\lambda_i \in s_{1:m}$, $\gamma_1 = \lambda_1$, and $\gamma_i = \lambda_i - \lambda_{i-1}$.
The $\lambda_{1:k}$ representation is the number of instances before every discretization edge whereas the $\gamma_{1:k}$ representation is the number of instances within each discretization interval. Both $\lambda_{1:k}$ and $\gamma_{1:k}$ representations will be used in the following content. The former makes the dynamic programming algorithm more succinct, and the latter makes equations more compact when calculating data instantiations. Table \ref{table:data_demo} and \ref{table:data_demo_2} demonstrate variable assignments for an example dataset.

\begin{table}[ht]
  \centering
  \caption{
    An example dataset used to demonstrate the variables defined in Section \ref{subsec:disc_policy}. Rows are assignments to continuous variable $\cX$ and discrete variables $A$ and $B$.
  }
  \scriptsize
\begin{tabular}{ccc}
\toprule
$\cX$ & $A$ & $B$ \\ \midrule
$1.1$ & $1$ & $3$ \\ 
$1.2$ & $1$ & $3$ \\ 
$1.2$ & $1$ & $4$ \\ 
$2.9$ & $2$ & $3$ \\ 
$2.9$ & $2$ & $4$ \\ 
$2.9$ & $1$ & $4$ \\ 
$3.5$ & $1$ & $4$ \\ 
$4.3$ & $1$ & $4$ \\ 
$4.3$ & $2$ & $3$ \\ 
$5.0$ & $2$ & $3$ \\ \bottomrule
\end{tabular}
  \label{table:data_demo}
  \bigskip
  \centering
  \caption{
    Assignments for variables defined in Section~\ref{subsec:disc_policy} according to the dataset in Table~\ref{table:data_demo}. Note that $\Lambda^{\text{ex}}_{\cX}$ is an example of a valid discretization policy. Discretization policies can be represented by their edges $\anglebrackets{}_{e}$, their sorted unique dataset indices $\anglebrackets{}_{\lambda}$, or by the interval between sorted unique dataset indices $\anglebrackets{}_{\gamma}$. That is to say, in the edge representation, $e_1 = 1.55$ and $e_2 = 4.65$. In the $\lambda_{1:k}$ representation, $\lambda_1 = 3$, $\lambda_2 = 9$, and $\lambda_3 = 10$. In the $\gamma_{1:k}$ representation, $\gamma_1 = 3$, $\gamma_2 = 6$, and $\gamma_3 = 10$. Furthermore, one can check that $\lambda_1 = \gamma_1 = 3$, $\lambda_2 - \lambda_1 = \gamma_2$, and $\lambda_3 - \lambda_2 = \gamma_2$.
  }
	\begin{tabular}{cc}
		\toprule
		Variable &  Value \\
		\midrule
		$n$ &  \num{10} \\
		$m$ &  \num{6} \\
		$\cx_{1:10}$ & $\curly{ 1.1, 1.2, 1.2, 2.9, 2.9, 2.9, 3.5, 4.3, 4.3, 5.0} $\\
		$\cu_{1:6} $ & $\curly{ 1.1,1.2,2.9,3.5,4.3,5.0}$\\
		$s_{1:6}$ & $\curly{1,3,6,7,9,10}$\\
		$\Lambda^{\text{ex}}_{\cX}$ & $\anglebrackets{1.55,4.65}_{e}$, $ \anglebrackets{3 ,9,10}_{\lambda}$, $\anglebrackets{3, 6, 1}_{\gamma} $\\
		\bottomrule
	\end{tabular}
  \label{table:data_demo_2}
\end{table}

\subsection{Bayesian Networks}

A Bayesian Network $B$ over $N$ random variables $X_{1:N}$ is defined by a directed acyclic graph $G$ containing one node for each variable $X_i$, and each node is associated with a conditional probability distribution $P(X_i \mid \Pa_{X_i})$ for the node given its parents in $G$. The edges in a Bayesian network represent probabilistic dependencies among nodes and encode the Markov property: each node $X_i$ is independent of its non-descendants given its parents $\Pa_{X_i}$ in $G$. The children of node $X_i$ are denoted $\Ch_{X_i}$.

When discussing the discretization of a particular continuous variable $\cX$, let $P_i$ be the $i$th parent of $\cX$, let $C_i$ be the $i$th child of $\cX$, and let $\bm{S}_i$ be the set of spouses of $\cX$ associated with the $i$th child. The parents, children, and spouses of a variable form its Markov blanket.


\section{Single Variable Discretization}
\label{sec:single_var}

This section covers the discretization of a single continuous variable $\cX$ in a Bayesian network where all other variables are discrete.
An optimal discretization policy for a dataset $D$ maximizes $P(\Lambda)\cdot P(D\mid \Lambda)$ for some prior $P(\Lambda)$ and likelihood $P(D\mid \Lambda)$.

\subsection{Priors and Objective Function}

Let $D_{\bm{Y}}$ be the subset of the training data corresponding to variable subset $\bm{Y}$.
Four principles for the optimal discretization policy enable the formulation of $P(\Lambda)$ and $P(D_{-\cX} \mid \Lambda)$, where $D_{-\cX}$ is the subset of the dataset for all variables but $\cX$, and the probability of the data associated with the target variable, $P(D_{\cX})$, is already captured in $P(\Lambda)$.
The four principles, which can be considered an extension of the priors in MODL \citep{Boulle_2006} and \citet{Lustgarten_2011} to Bayesian networks, are:

\begin{enumerate}
\item The prior probability of a discretization edge between two consecutive unique values $\cu_i$ and $\cu_{i+1}$ is proportional to their difference:
  \begin{equation}
  1 - \exp \paren{- L \cdot \frac{\cu_{i+1} - \cu_i}{\cu_m - \cu_1}}\text{,}
  \end{equation}

where larger values of $L$ encourage more discretization intervals.
This prior encourages edges between well-separated values over edges between closely packed samples.
The experiments in this paper set $L$ to the largest number of intervals among discrete variables in $\cX$'s Markov blanket, which encourages Bayesian networks with consistent interval counts.
Higher values of $L$ can be used when the continuous variable requires more discretization intervals than its Markov blanket.

\item For a given discretization interval, every distribution over the parents of $\cX$ is equiprobable.
\item For each pair $\anglebrackets{C_i,\bm{S}_i}$ and a given discretization interval, every distribution over $C_i$ given an instance of $\bm{S}_i$ is equiprobable.
\item The distributions over $\cX$ given each child, parent, and spouse instantiation are independent.
\end{enumerate}

From the first principle one obtains the prior over the discretization policy:
\begin{small}
\begin{equation}
  \label{eq:p_M}
  P(\Lambda) \propto \prod_{i=1}^{k-1}
    \brock{
      1 - \exp\paren{
        - L \cdot \frac{
                         \cx_{\lambda_i +1} - \cx_{\lambda_i}
                       }{
                         \cx_n - \cx_1
                        }
      }
    }
    \prod_{i=1}^{k}
	\brock{
    \exp\paren{
      -L \cdot \frac{\cx_{\lambda_{i}} - \cx_{\lambda_{i-1} + 1}}{\cx_n - \cx_1}}
    }\text{.}
\end{equation}
\end{small}

The likelihood term ${P(D_{-\cX} \mid \Lambda)}$ for a Bayesian network graph structure factors according to $\cX$'s Markov blanket:
\begin{equation}
  \label{eq:p_D_given_M}
  P\paren{D_{-\cX} \mid \Lambda} \propto P\paren{D_{\Pa_\cX} \mid \Lambda} \cdot \prod_{i} P\paren{D_{C_i} \mid \Lambda, D_{\bm{S}_i}}\text{.}
\end{equation}

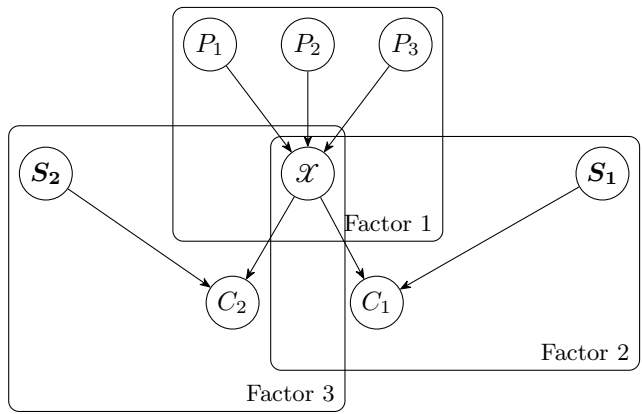
\begin{figure}[ht]
  \centering
%
%
%
%


\begin{tikzpicture}[>={Stealth[round]}
      ]

  \node[latent]                               (c1) {$C_2$};
  \node[latent, right= 1.2 cm of c1]            (c2) {$C_1$};
  \node[latent, above=of c1, xshift= 1 cm] (y)  {$\cX$};
  \node[latent, above=of y, xshift=-1.3cm] (p1) {$P_1$};
  \node[latent, above=of y, xshift=0cm]  (p2) {$P_2$};
  \node[latent, above=of y, xshift=1.3cm]  (p3) {$P_3$};
  \node[latent, right= 3.2cm of y]            (s1) {$\boldsymbol{S_1}$};
  \node[latent, right=-4.2cm of y]            (s2) {$\boldsymbol{S_2}$};

  \edge {p1,p2,p3} {y} ; %
  \edge {y,s1}{c2} ;
  \edge {y,s2}{c1};

  \plate {} {(p1)(p2)(p3)(y)} {$\text{Factor 1}$} ;
  \plate {yx} {(y)(s1)(c2)} {$\text{Factor 2}$};
  \plate {} {(y)(s2)(c1)(yx.north west)(yx.south west)} {$\text{Factor 3}$} ;

\end{tikzpicture}

  \caption{Factorization of $P(D_{-\cX} \mid \Lambda)$.}
  \label{fig:example_factorization}
\end{figure}

For example, in Figure~\ref{fig:example_factorization}, $P(D_{-\cX} \mid \Lambda)$ factors according to
\begin{equation}
  P\paren{D_{P_1,P_2,P_3} \mid \Lambda} \cdot P\paren{ D_{ C_1 } \mid \Lambda, D_{\bm{S}_1} } \cdot P\paren{D_{C_2} \mid \Lambda, D_{\bm{S}_2}}\text{.}
\end{equation}

The concept behind the factorization is also the motivation behind forward sampling in a Bayesian network.
The parents of $\cX$ are independent of the children given $\cX$.
The parents are not necessarily individually independent, and thus the parental term $P(D_{\Pa_{\cX}} \mid \Lambda)$ cannot be factored further.
The children of $\cX$ are similarly independent given $\cX$ and the corresponding spouses, leading to their factored product $\prod_{i} P(D_{C_i} \mid \Lambda, D_{\bm{S}_i})$.
Each component in the decomposition can be evaluated given a discretization policy and a dataset.

\subsubsection{Evaluation of $P\paren{D_{\Pa_{\cX}} \mid \Lambda}$}

Let $J_\text{P}$ be the number of instantiations of the parents of $\cX$, and let $n^{(\text{P})}_{i,j}$ be the number of instances of $\cX$ within the $i$th discretization interval of $\Lambda$ given the $j$th parental instantiation.
Note that $\gamma_i = \sum_j n^{(\text{P})}_{i,j}$.
It follows that
\begin{equation}
  \label{eq:likelihood_one}
  P\paren{D_{\Pa_{\cX}} \mid \Lambda} = \overbrace{\prod_{i=1}^k}^{\text{Principle 4}}
    \overbrace{\rule{0pt}{2em}
      \frac{1}{{{\gamma_i + J_\text{P} - 1}\choose{J_\text{P} - 1}}}
      \frac{1}{
        \frac{
          {\gamma_i}!
        }{
          {n^{(\text{P})}_{i,1}}! \; {n^{(\text{P})}_{i,2}}! \; \cdots \; {n^{(\text{P})}_{i,J_\text{P}}}!
        }
      }
    }^{\text{Principle 2}}\text{.}
\end{equation}

Both factors on the right hand side come from the second principle: all distributions of values of the parents of $\cX$ in a given interval are equiprobable.
The fourth principle is that the distributions for each interval are independent, so the factors can be multiplied together. Note that the combinatorics of factors ${{\gamma_i + J_\text{P} - 1}\choose{J_\text{P} - 1}}$ and ${
	{\gamma_i}!}/{{n^{(\text{P})}_{i,1}}! \cdots \; {n^{(\text{P})}_{i,J_\text{P}}}!}$ are explained in the MODL paper~\citep{Boulle_2006}.

\subsubsection{Evaluation of $P(D_{C_j} \mid \Lambda, D_{\bm{S}_j})$}
Let $J_{C_j}$ be the number of instantiations of the $j$th child of $\cX$, let $J_{\bm{S}}^{(j)}$ be the number of instantiations of the $j$th spouse set $\bm{S}_j$, and let $n^{(j)}_{i,m,\ell}$ be the number of instances of $\cX$ in the $i$th discretization interval of $\Lambda$ given the $m$th instantiation of $C_j$ and the $\ell$th instantiation of $\bm{S}_j$.
Let ${n^{(j)}_{i,\ell} = \sum_{m=1}^{J_j} n^{(j)}_{i,m,\ell}}$.
Note that ${\gamma_i = \sum_{\ell} n^{(j)}_{i,\ell}}$ for all $j$.
It follows that
\begin{equation}
  \label{eq:likelihood_two}
  P(D_{C_j} \mid \Lambda, D_{\bm{S}_j}) =
  \overbrace{\prod_{i=1}^{k} \prod_{\ell=1}^{J_{\bm{S}}^{(j)}}}^{\text{Principle 4}}
    \overbrace{\rule{0pt}{2.3em}
      {
        \frac{1}{
          {{n^{(j)}_{i,\ell} + J_{C_j} - 1}\choose{J_{C_j}-1}}}
        }
        {\frac{
          1
        }{
          \frac{
            {n^{(j)}_{i,\ell}}!
          }{
            {n^{(j)}_{i,1,\ell}!} \; {n^{(j)}_{i,2,\ell}!} \; \cdots \; {n^{(j)}_{i,J_{C_j},\ell}!}
          }
        }
      }
    }^{\text{Principle 3}}
    \text{.}
\end{equation}

Both factors on the right hand side come from the third principle: all distributions of values of $C_j$ in a given interval and with a given value of $S_j$ are equiprobable.
According to the fourth prior, these distributions are independent from each other, and one can thus take their product.
If $\bm{S}_j = \emptyset$, then Equation~\ref{eq:likelihood_two} is equivalent to
\begin{equation}
  \label{eq:likelihood_three}
  P(D_{C_j} \mid \Lambda, \bm{S}_j = \emptyset) =
  \overbrace{\prod_{i=1}^{k}}^{\text{Principle 4}}
  \overbrace{\rule{0pt}{2.3em}{
    {1}\over{
      {\gamma_i + J_{C_j} - 1}\choose{J_{C_j}-1}}
    }
    {{1}\over{
      {{\gamma_i}!} \over {
        {n^{(j)}_{i,1,\emptyset}!} \; {n^{(j)}_{i,2,\emptyset}!} \; \cdots \; {n^{(j)}_{i,J_{C_j},\emptyset} !}
      }
    }
  }}^{\text{Principle 3}}
  \text{,}
\end{equation}
where $n^{(j)}_{i,m,\emptyset}$ is the number of instances of $\cX$ in the $i$th discretization interval of $\Lambda$ given the $m$th instantiation of $C_j$.

The objective function can be formulated given equations~\ref{eq:p_M},~\ref{eq:p_D_given_M},~\ref{eq:likelihood_one},~\ref{eq:likelihood_two} and~\ref{eq:likelihood_three}.
The log-inverse of $P(\Lambda) \cdot P(D_{-\cX} \mid \Lambda)$ is minimized for computational convenience:
\begin{equation}
\label{eq:opt_prob}
\begin{aligned}
  & \sum_{i=1}^{k-1}
   - \ln
    \paren{
      1 - \exp
      \paren{
        - L \cdot  \frac{
                         \cx_{\lambda_i +1} - \cx_{\lambda_i}
                       }{
                         \cx_n - \cx_1
                        }
      }
    }
      + \sum_{i=1}^{k}
      L \cdot \frac{\cx_{\lambda_{i}} - \cx_{\lambda_{i-1} + 1}}{\cx_n - \cx_1} + \\
  & \sum_{i=1}^{k} \brock{
  \ln{{\gamma_i + J_\text{P} - 1}\choose{J_\text{P} - 1}}
  +{ \ln \paren{
        \frac{
          {\gamma_i}!
        }{
          {n^{(\text{P})}_{i,1}}! \; {n^{(\text{P})}_{i,2}}! \; \cdots \; {n^{(\text{P})}_{i,J_\text{P}}}!
        }
      }}
  } + \\
  & \sum_{j=1}^{n_c} \sum_{i=1}^k \sum_{\ell=1}^{J^{(j)}_{\bm{S}}} \brock{
  { \ln
          {{n^{(j)}_{i,\ell} + J_{C_j} - 1}\choose{J_{C_j}-1}}}
    +
    \ln \paren{ {
          \frac{
            {n^{(j)}_{i,\ell}}!
          }{
            {n^{(j)}_{i,1,\ell}!} \; {n^{(j)}_{i,2,\ell}!} \; \cdots \; {n^{(j)}_{i,J_{C_j},\ell}!}
          }
        }
        } }
\end{aligned}
\end{equation}


\subsection{Single-Variable Discretization Algorithm}
\label{subsec:algo}

The procedure used to minimize the objective function involves dynamic programming.
Note that because the objective function is cumulative over intervals, if a partition $\Lambda = \anglebrackets{\gamma_1, \gamma_2, \ldots, \gamma_k}$ of $\cX$ is an optimal discretization policy, then any subinterval is optimal for the corresponding subproblem.
It follows that dynamic programming can be used to solve the optimization problem exactly.

Precomputation reduces runtime.
Hence, $h(u,v)$ is computed first for each interval $\gamma_q = \gamma_q^{(u,v)}$ starting from $x_{u}$ to $x_{v}$ for all $u$, $v$ satisfying $u \leq v$:
\begin{small}
  \begin{equation}
  \label{eqn:h_function}
  \begin{aligned}
  h(u,v) &=  \ln {{\gamma_{q} + J_\text{P} - 1}\choose{J_\text{P}-1}} + \ln \left( { {{\gamma_q}!}\over{ {n^{(\text{P})}_{q,1} !} {n^{(\text{P})}_{q,2} !} \cdots {n^{(\text{P})}_{q,J_p} !}} } \right) \\
  & + \sum_{j=1}^{n_c} \sum_{\ell=1}^{J^{(j)}_{\bm{S}}} \brock{
    { \ln
            {{n^{(j)}_{i,\ell} + J_{C_j} - 1}\choose{J_{C_j}-1}}}
      +
      \ln \paren{ {
            \frac{
              {n^{(j)}_{i,\ell}}!
            }{
              {n^{(j)}_{i,1,\ell}!} \; {n^{(j)}_{i,2,\ell}!} \; \cdots \; {n^{(j)}_{i,J_{C_j},\ell}!}
            }
          }
          } }
  \end{aligned}
  \end{equation}
\end{small}

The calculation of $h(u,v)$ for all $u \leq v$ has a $\bigo \paren{ n_c \cdot {L}^{n_s} \cdot n^2 + {L}^{n_p} \cdot n^2 }$ runtime, where $n_c$ and $n_p$ are the number of child and parent variables respectively, $L$ is the largest cardinality of variables in $X$'s Markov blanket, and $n_s = \text{max}_j  |\Pa_{C_j}|$. The $n^2$ factor in the runtime comes from the fact that both $x_u$ and $x_v$ have to run over all possible values of $\cX$, which has $n$ instantiations.

The optimization problem over Equation~\ref{eq:opt_prob} can now be solved.
The dynamic programming procedure is shown in Algorithm~\ref{alg:disc_one}.
It takes three inputs: $\cX$, the continuous variable; $D$, the joint data instances over all variables sorted in ascending order according to $D_X$; and $G$, the network structure.
The runtime of Algorithm~\ref{alg:disc_one} is also $\bigo \paren{ n_c \cdot {L}^{n_s} \cdot n^2 + {L}^{n_p} \cdot n^2 }$ because the runtime of the dynamic programming procedure is less than the runtime for computing $h(u,v)$.
As will be discussed in Section~\ref{sec:experiments}, the MDL discretization method has a runtime of $\bigo\paren{ n^3 + \paren{ n_c \cdot {L}^{n_s} + {L}^{n_p}} \cdot n^2 }$, which includes an extra $\bigo\paren{ n^3}$ term.
The Bayesian discretization method is quadratic in the sample count $n$, whereas the MDL discretization method is cubic.

Algorithm~\ref{alg:disc_one} is guaranteed to be optimal.
For faster methods with suboptimal results, see \citet{Boulle_2006}.

\begin{algorithm}
  \caption{Discretization of one continuous variable in a Bayesian network}
  \label{alg:disc_one}
  \begin{algorithmic}[5]
    \Function{DiscretizeOne}{$D$, $G$, $\cX$}
      \State $H \leftarrow$ an $n \times n$ matrix such that $H[u,v] = h(u,v)$; can be precomputed
      \State $L \leftarrow$ the largest cardinality over all discrete variables in the Markov blanket of $\cX$
      \State $S[i] \leftarrow$ the optimal objective value computed over samples $1$ to $s_i$
      \State $\Lambda[i] \leftarrow$ the discretization policy for the subproblem over samples $1$ to $s_i$
      \State $W[i]  \leftarrow - \ln\brock{1 - {\exp\paren{- L \cdot{ {{\cx_{s_i+1} - \cx_{s_i}}\over{\cu_m - \cu_1}}}}}}$ for $i \in [1,m-1]$ and $W[m] \leftarrow 0$
      \For {$v \leftarrow 1$ to $m$}
        \If {$v = 1$}
          \State $S[v] \leftarrow H \paren{1,s_v} + L[v]$
          \State $\Lambda[v] \leftarrow \curly{({\cu_v + \cu_{v+1}}) / 2}$
        \Else
          \State $\hat{S}, \hat{u} \leftarrow \infty, 0$
          \State $\text{DiscEdge} \leftarrow \infty$
          \For {$u \leftarrow 1$ to $v$}
            \If {$u = v$}
              \State $\tilde{S} \leftarrow W[v] + H \paren{1, s_v} +  {L \cdot {\frac{\cu_{v} - \cu_1}{\cu_m - \cu_1}}}$
            \Else
              \State $\tilde{S} \leftarrow W[v] + H \paren{s_u+1, s_v} +  {L \cdot {\frac{\cu_{v} - \cu_{u + 1}}{\cu_m - \cu_1}}} + S[u]$
            \EndIf
            \If {$\tilde{S} < \hat{S}$}
              \State $\hat{S}, \hat{u} \leftarrow \tilde{S}, u$
              \State $\text{DiscEdge} \leftarrow ({\cx_{s_u} + \cx_{s_u+1}}) / 2$
            \EndIf
          \EndFor
          \State $S[v] \leftarrow \hat{S}$
          \State $\Lambda[v] \leftarrow \Lambda[\hat{u}] \cup \{ \text{DiscEdge} \}$
        \EndIf
      \EndFor
      \State \Return $\Lambda[m]$
    \EndFunction
  \end{algorithmic}
\end{algorithm}

\section{Multi-variable Discretization}
\label{sec:multi_var}

The single-variable discretization method can be extended to Bayesian networks with multiple continuous variables by iteratively discretizing individual variables.
The discretization process for a single variable requires that all other variables be discrete.
The iterative approach uses an initial discretization policy in order to start the process which assigns $k$ equal-width intervals to each continuous variable, where $k$ is the largest number of intervals of initially discrete variables in the network.

\begin{figure}[ht]
 \centering
%
%
%
%


\begin{tikzpicture}[
      >={Stealth[round]}
      ]

  \node[latent]                               (c1) {$C_1$};
  \node[latent, right= 1.7 cm of c1]            (c2) {$C_2$};
  \node[latent, above=of c1, xshift= 1.2 cm] (y)  {$X$};
  \node[obs, above=of c2, xshift= 1.2 cm] (s)  {$S_2$};
  \node[latent, above=of y, xshift=-1.2cm] (p1) {$P_1$};
  \node[latent, above=of y, xshift=1.2cm]  (p2) {$P_2$};

  \edge {p1,p2} {y} ; %
  \edge {y} {c2} ;
  \edge {y}{c1};
  \edge {s}{c2};


\end{tikzpicture}

  \caption{An example network.}
   \label{fig:example_networks}
\end{figure}
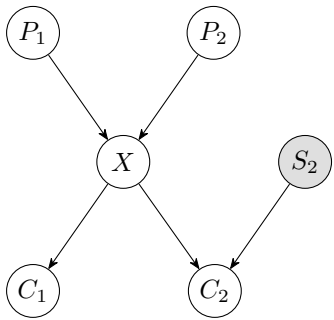

After the initial discretization, the one-variable discretization method is iteratively applied over each continuous variable in reverse topological order, from the leaves to the root.
Reverse topological order has the advantage of relying on fewer initial discretizations of the continuous variables during the first pass.
For example, in the network of Figure~\ref{fig:example_networks}, if $S_2$ is the only discrete variable, then the discretization of $P_1$ involves both $P_2$ and $X$, whereas the discretization of $C_1$ only involves $X$.

The algorithm is terminated when the number of discretization intervals and their associated edges converge for all variables, and a maximum number of complete passes is enforced to prevent infinite iterations when convergence does not occur.
The algorithm typically converges within a few passes when tested on real-world data.

The pseudocode for the multi-variable discretization procedure is shown in Algorithm~\ref{alg:disc_two}.
It requires four inputs: $D$, a dataset of samples from the joint distribution; $G$, the fixed network structure; $\bm{\cX}$, the set of all continuous variables in reverse topological order, and $\maxpasses$, an upper bound on the number of complete passes.

\begin{algorithm}
  \caption{Discretization of multiple continuous variables}
  \label{alg:disc_two}
  \begin{algorithmic}[5]
  \Function{DiscretizeAll}{$D$, $G$, $\bm{\cX}$, $\maxpasses$}
    \State $\discset \leftarrow$ the discretization policies for each $\cX$ in $\bm{\cX}$
    \State $D^* \leftarrow $ the dataset $D$ discretized according to $\discset$
    \State $k \leftarrow \textsc{Max}\paren{\curly{ |X| \text{ s.t. } X \text{ does not have corresponding } \cX \text{ in } \bm{\cX} }}$
    \For {$\paren{\cX, X}$ such that $\cX \in \bm{\cX}$ and $X$ is the discretized version of $\cX$}
      \State $\Lambda_{\cX} \leftarrow$ equal-width discretization of $\cX$ with $k$ intervals
      \State $D^*_{X} \leftarrow \Lambda_{\cX} \paren{D_{\cX}}$
    \EndFor
    \State $n_\text{cycle} \leftarrow 0$
    \While {$\discset$ has not converged \textbf{and} $n_\text{cycle} \leq \maxpasses$}
      \State increment {$n_\text{cycle}$}
      \For {$\paren{\cX, X}$ such that $\cX \in \bm{\cX}$}
        \State {$D^*_X \leftarrow D_{\cX}$}
        \State {$\Lambda_{\cX} \leftarrow \textsc{DiscretizeOne}(D^*, G, \cX)$}
        \State {$D^*_X \leftarrow \Lambda_{\cX}\paren{D_{\cX}}$}
      \EndFor
    \EndWhile
    \State \Return $\discset$
  \EndFunction
  \end{algorithmic}
\end{algorithm}

\section{Combining Discretization with Structure Learning}
\label{sec:structure_learning}

It is often necessary to infer the structure of a Bayesian network from data.
Three common approaches to Bayesian network structure learning are constraint-based, score-based, and Bayesian model averaging \citep{PGM_2009}.
This work uses the K2 structure learning algorithm \citep{cooper1992bayesian}, a frequently used score-based structure learning method.
Score-based structure learning methods over discrete variables commonly evaluate candidate structures according to their likelihood against a dataset $D$.
In practice, one maximizes the log-likelihood, also known as the Bayesian score, $\ln P(G\mid D)$~\citep{cooper1992bayesian}.

For a Bayesian network over $N$ discrete and discretized variables $X_{1:N}$, $r_i$ represents the number of instantiations of $X_i$, and $q_i$ represents the number of instantiations of the parents of $X_i$.
If $X_i$ has no parents, then $q_i = 1$.
The Bayesian score is:
\begin{small}
\begin{equation}
\label{eq:BayesianScore}
  \ln P(G\mid D) = \ln P(G) + \sum_{i=1}^N \sum_{j=1}^{q_i} \ln \paren{
    \frac{
      \mathrm{\Gamma}\paren{\alpha_{ij}^{(0)}}
    }{
      \mathrm{\Gamma}\paren{\alpha_{ij}^{(0)} + \beta_{ij}^{(0)}}
    }
  } + \sum_{k=1}^{r_i} \ln \paren{
    \frac{
      \mathrm{\Gamma}\paren{\alpha_{ij}^{(k)} + \beta_{ij}^{(k)}}
    }{
      \mathrm{\Gamma}\paren{\alpha_{ij}^{(k)}}
    }
  }\text{,}
\end{equation}
\end{small}
where $\mathrm{\Gamma}$ is the gamma function, $\alpha_{ij}^{(k)}$ is a Dirichlet parameter and $\beta_{ij}^{(k)}$ is the observed sample count for the $k$th instantiation of $X_i$ and the $j$th instantiation of $\Pa_{X_i}$.
In the equation above,
\begin{equation}
\alpha_{ij}^{(0)} = \sum_{k=1}^{r_i} \alpha_{ij}^{(k)} \quad \qquad \beta_{ij}^{(0)} = \sum_{k=1}^{r_i} \beta_{ij}^{(k)}\text{.}
\end{equation}

The space of acyclic graphs is superexponential in the number of nodes; it is common to rely on heuristic search strategies \citep{PGM_2009}.
The K2 algorithm assumes a given topological ordering of variables and greedily adds parents to nodes to maximally increase the Bayesian score.
A fixed ordering ensures acyclicity but does not guarantee a globally optimal network structure.
K2 is typically run multiple times with different topological orderings, and the network with the highest likelihood is retained.

Traditional Bayesian structure learning algorithms require discretized data, whereas the proposed discretization algorithm requires a known network structure.
The discretization methods can be combined with the K2 structure learning algorithm \citep{cooper1992bayesian} to simultaneously perform Bayesian structure learning and discretization of continuous variables.

The proposed algorithm alternates between K2 structure learning and discretization.
The dataset is initially discretized as was described in Section~\ref{sec:multi_var}, and K2 is run to obtain an initial network structure.
The affected continuous variables are rediscretized every time an edge is added by K2.
The resulting discretization policies are used to update the discretized dataset, and the next step of the K2 algorithm is executed.
This cycle is repeated until the K2 algorithm converges.

This procedure is given in Algorithm~\ref{alg:structure_learn}, and takes five inputs: $D$, a dataset of samples from the joint distribution; $\bm{\cX}$, the set of all continuous variables; $\texttt{order}$, a permutation of the variables in $D$; $\hat{n}_\text{parent}$, an upper bound on the number of parents per node; and $\maxpasses$, an upper bound on the number of complete passes.
The upper bound on the number of parents is common practice, and is used to prevent computing conditional distributions with excessively large parameter sets.
The function $g$ in Algorithm~\ref{alg:structure_learn} computes a component of the Bayesian score (Equation~\ref{eq:BayesianScore}):
\begin{equation}
  \label{eq:B_Score_one}
  g\paren{X_i, \Pa_{X_i}} =  \sum_{j=1}^{q_i} \ln\paren{
    \frac{
      \mathrm{\Gamma}\paren{\alpha_{ij}^{(0)}}
    }{
      \mathrm{\Gamma}\paren{\alpha_{ij}^{(0)} + \beta_{ij}^{(0)}}
    }
  } + \sum_{k=1}^{r_i} \ln \paren{
    \frac{
      \mathrm{\Gamma}\paren{\alpha_{ij}^{(k)} + \beta_{ij}^{(k)}}
    }{
      \mathrm{\Gamma}\paren{\alpha_{ij}^{(k)}}
    }
  }\text{.}
\end{equation}

\begin{algorithm}
  \caption{Learning a discrete-valued Bayesian network}
  \label{alg:structure_learn}
  \begin{algorithmic}[5]
  \Function{LearnDVBN}{$D$, $\bm{\cX}$, $\texttt{order}$, $\hat{n}_\text{parent}$, $\maxpasses$}
    \State $N \leftarrow$ the number of variables in the Bayesian network
    \State $k \leftarrow \textsc{Max} \{ \|v\|, v\notin C\}$
    \State $\discset \leftarrow$ the discretization policies for each $\cX$ in $\bm{\cX}$
    \State $D^* \leftarrow $ the dataset $D$ discretized according to $\discset$
    \State $G \leftarrow$ the initially edgeless graph structure
	\For {$\paren{\cX, X}$ such that $\cX \in \bm{\cX}$ and $X$ is the discretized version of $\cX$}
      \State $\Lambda_{\cX} \leftarrow$ equal-width discretization of $\cX$ with $k$ intervals
      \State $D^*_{X} \leftarrow \Lambda_{\cX} \paren{D_{\cX}}$
    \EndFor
    \For {$i \leftarrow 1$ to $N$}
      \State $P_\text{old} \leftarrow g(X_i,\Pa_{X_i})$ \quad (Equation~\ref{eq:B_Score_one})
      \State OKToProceed $\leftarrow$ \textbf{true}
      \While {OKToProceed \textbf{and} $\|\Pa_{X_i}\| < \hat{n}_\text{parent}$}
        \State $Y \leftarrow$ an element from the set $\texttt{order}[1:i] \backslash \paren{\Pa_X}$
        \State $P_\text{new} \leftarrow \paren{g(X_i,\Pa_{X_i}} \cup \paren{Y})$
        \If {$P_\text{new} > P_\text{old}$}
          \State $P_\text{old} \leftarrow P_\text{new}$
          \State $\Pa_{X_i} \leftarrow \paren{\Pa_{X_i}} \cup \paren{Y} $
          \State $\bm{\cX '} \leftarrow$ $\bm{\cX}$ sorted in reverse-topological order given $G$
          \State $\Lambda \leftarrow$ \textsc{DiscretizeAll}({$D$, $G$, $\bm{\cX '}$, $\maxpasses$}) \quad (see Algorithm~\ref{alg:disc_two})
          \State $D^* \leftarrow \Lambda \paren{D_{\bm{\cX }}}$
        \Else
          \State OKToProceed $\leftarrow$ \textbf{false}
        \EndIf
      \EndWhile
    \EndFor
    \State \Return $G$, $\Lambda$
  \EndFunction
  \end{algorithmic}
\end{algorithm}

It is common practice to run K2 multiple times with different variable permutations and to then choose the structure with the highest score.
As such, Algorithm~\ref{alg:structure_learn} is run multiple times with different variable permutations and the discretized Bayesian network with the highest score is retained.


\section{Experiments}
\label{sec:experiments}

This section describes experiments conducted to evaluate the Bayesian discretization method.
All experiments were run on datasets from the publicly available University of California, Irvine machine learning repository \citep{Lichman_2013}.
Variables are labeled alphabetically in the order given on the dataset information webpage.
In the figures that follow, shaded nodes correspond to initially discrete variables and the subscripts indicate the number of discrete instantiations.

Two experiments were conducted on each dataset.
The first experiment compares the performance of the Bayesian and MDL discretization methods on a known Bayesian network structure.
The structure was obtained by discretizing each continuous variable into $k$ uniform-width intervals, where $k$ is the median number of instantiations of the discrete variables, and using the structure with the highest Bayesian score from \num{1000} runs of the K2 algorithm with random topological orderings.
The second experiment compares the same methods applied when simultaneously discretizing and learning network structure.

The discretizations are compared using the mean cross validated log-likelihood of the data $D$ given the Bayesian network $B$ and discretization policies $\discset$.
Note that all log-likelihoods shown have been normalized by the number of data samples.
The log-likelihood has two components,
\begin{equation}
\ln p(D\mid B, \discset) = \ln P(D^*\mid B) + \ln p(D\mid \discset, D^*)\text{,}
\end{equation}

\noindent
where $D$ is the original test dataset and $D^*$ is the test dataset discretized according to $\discset$.

The log-likelihood of the discretized dataset given the Bayesian network is
\begin{equation}
P(D^* \mid B) = \prod_{i=1}^N \prod_{j=1}^{q_i} \prod_{k=1}^{r_i}
\paren{
  \frac{
    \alpha_{ij}^{(k)} + \beta_{ij}^{(k)}
  }{
    \alpha_{ij}^{(0)} + \beta_{ij}^{(0)}
    }
}^{\beta_{ij}^{*(k)}}\text{,}
\end{equation}

\noindent
where $\bm{\alpha}$ and $\bm{\beta}$ are the Dirichlet prior counts and observed counts in the training set for $B$, and $\bm{\beta}^*$ is the set of observed counts in the test set $D^*$.
All experiments used a uniform Dirichlet prior of $\alpha_{ijk} = 1$ for all $i$, $j$, and $k$.

The log-likelihood of the original dataset given the discrete dataset is
\begin{equation}
  \ln p\paren{D\mid \discset, D^*} = \sum_{i=1}^N
  1_{\curly{ \cX_i \in \bm{\cX }}}\cdot \sum_{k=1}^{r_i}  \sum_{j=1}^{q_i} \beta_{ij}^{*(k)} \ln \paren{
    \frac{
      1
    }{
      e^{\Lambda_{\cX_i}}_{k} - e^{\Lambda_{\cX_i}}_{k-1}
    }
  }\text{.}
\end{equation}
Note that $e^{\Lambda_{\cX_i}}_{0} = \textsc{Min}\paren{D_{\cX_i}}$ and $e^{\Lambda_{\cX_i}}_{r_i} = \textsc{Max}\paren{D_{\cX_i}}$.
\noindent
The mean cross-validated log-likelihood is the mean log-likelihood on the witheld dataset among cross-validation folds, and acts as an estimate of generalization error.
Ten folds were used in each experiment.

The method for computing the MDL discretization policy is similar to the method for the Bayesian method.
For a Bayesian network with a single continuous variable $\cX$, the MDL objective function is
\begin{equation}
  \label{eqn:MDL}
  \begin{aligned}
  \frac{1}{2} \ln(n) \left\lbrace  \lvert \Pa_{\cX} \rvert (\lvert X \rvert - 1) +
   {\sum_{j,\cX \in \Pa_{X_j}}} \lvert \Pa_{X_j} \rvert (\lvert X_j \rvert - 1) \right\rbrace + \ln(|X|) \\
   + (m-1) H \paren{\frac{|X| - 1}{m -1}} -n \cdot \left[ I(X,\Pa_{\cX}) + {\sum_{j,\cX \in \Pa_{X_j}}} I(X_j, \Pa_{X_j}) \right]\text{,}
  \end{aligned}
\end{equation}
where $I(A,B)$ is the mutual information between two discrete variable sets $A$ and $B$, ${H(p) = -p \ln(p) - (1-p) \ln(1-p)}$, and $X$ is the discretized version of $\cX$.

The global minimum of Equation~\ref{eqn:MDL} can be found using dynamic programming.
For a given Bayesian network, the first three terms only depend on the number of discretization intervals $|X|$ and the fourth term is cumulative over the intervals.
Hence, if ${\Lambda = \anglebrackets{\gamma_1, \gamma_2, \ldots, \gamma_k}}$ is an MDL optimal discretization policy with $k$ intervals, then ${\Lambda' = \anglebrackets{\gamma_1, \gamma_2, \ldots, \gamma_{k-1}}}$ is an MDL optimal discretization policy for the corresponding subproblem with $k-1$ intervals.
This procedure takes runtime $\bigo \paren{ n^3 + \paren{ n_c \cdot {L}^{n_s}  + {L}^{n_p}} \cdot n^2 }$.
All variables are defined in Section \ref{subsec:algo}.
Therefore, the runtime of the proposed method in this paper is less than the MDL discretization method by $\bigo \paren{n^3}$, since the former has a more efficient form of dynamical programming.
There also exist faster methods of MDL discretization, but they lead to suboptimal results \citep{Friedman_1996}.

Note that the preliminary discretization and the discretization order of variables can be arbitrary for the MDL discretization method, as stated in the original work \citep{Friedman_1996}. In order to make a fair comparison between the Bayesian and MDL method, the preliminary discretization and the discretization order follows the same procedure as in Section \ref{sec:multi_var}.

\subsection{Dataset 1: Auto MPG}
\label{subsec:auto}

The Auto MPG dataset contains variables related to the fuel consumption of automobiles in urban driving.
The dataset has \num{392} samples over eight variables, not including six instances with missing data.
Three variables are discrete: $B$, $G$, and $H$, with \num{5}, \num{13}, and \num{3} instantiations respectively.

\subsubsection{Discretization with Fixed Structure}
\label{subsubsec:auto_exp1}

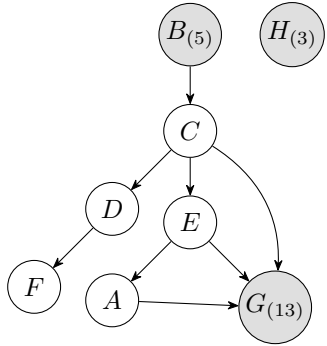
\begin{figure}[ht]
  \centering
%
%
%
%


\begin{tikzpicture}[
      >={Stealth[round]}
      ]

      \node[obs]                 (c2) {$B_{{(5)}}$};
      \node[latent, below=0.5 of c2]   (c3){$C$};
      \node[latent, below=0.5 of c3]   (c5){$E$};
      \node[latent, below left = 0.75 of c3] (c4){$D$};
      \node[latent, below left = 0.75 of c4] (c6){$F$};
      \node[latent, below left = 0.75 of c5] (c1){$A$};
      \node[obs, below right = 0.75 of c5] (c7) {$G_{(13)}$};
      \node[obs, right = 0.5 of c2] (c8){$H_{{(3)}}$};

  \edge {c2}{c3};
  \edge {c3}{c4,c5};
  \draw[->] (c3) to[bend left](c7);
  \edge {c4}{c6};
  \edge {c5}{c1,c7};
  \edge {c1}{c7};

\end{tikzpicture}

   \caption{Bayesian network structure obtained from running K2 on the initially uniformly discretized Auto MPG dataset.}
  \label{fig:auto_graph_1}
\end{figure}

The Bayesian and MDL discretization methods were tested on the Auto MPG data using the network shown in Figure~\ref{fig:auto_graph_1}.
This structure was obtained by initially discretizing each continuous variable into five uniform-width intervals, where five is the median cardinality of the discrete variables, and then taking the structure with highest likelihood from \num{1000} runs of K2.

Table~\ref{table:auto_disc_table_1} lists the discretization edges and mean log-likelihoods under \num{10}-fold cross validation of the discrete Bayesian network resulting from the two discretization methods.
The MDL method does not produce any discretization edges, assigning one continuous interval to each continuous variable, and produces the result with the lower likelihood.
The reason why MDL produces fewer discretization edges is discussed in Section~\ref{subsec:discuss_exp}.
Two examples that MDL produces comparable results to the Bayesian method are given in Appendix B.

\begin{table}[ht]
  \centering
  \caption{
    Discretization result of the Auto MPG dataset with fixed structure from Figure~\ref{fig:auto_graph_1}.
    The first five rows list the discretization edges and the last row lists the mean cross-validated log-likelihood; positive values are better.
  }
  \scriptsize
\begin{tabular}{@{}ccc@{}}
    \toprule
    Variable & Bayesian & MDL \\
    \midrule
    $A$ & \numlist[list-final-separator = {, }]{15.25; 17.65; 20.90; 25.65; 28.90} & - \\
    $C$ & \numlist[list-final-separator = {, }]{70.5; 93.5; 109.0; 159.5; 259.0; 284.5} & - \\
    $D$ & \numlist[list-final-separator = {, }]{71.5;99.0;127.0} & - \\
    $E$ & \numlist[list-final-separator = {, }]{2115;2481;2960;3658} & - \\
    $F$ & \numlist[list-final-separator = {, }]{12.35;13.75;16.05;22.85} & - \\
    \midrule
    Log-Likelihood & \num{-26.04} & \num{-30.40} \\
    \bottomrule
\end{tabular}
  \label{table:auto_disc_table_1}
\end{table}

\begin{figure}[ht]
  \centering
  \input{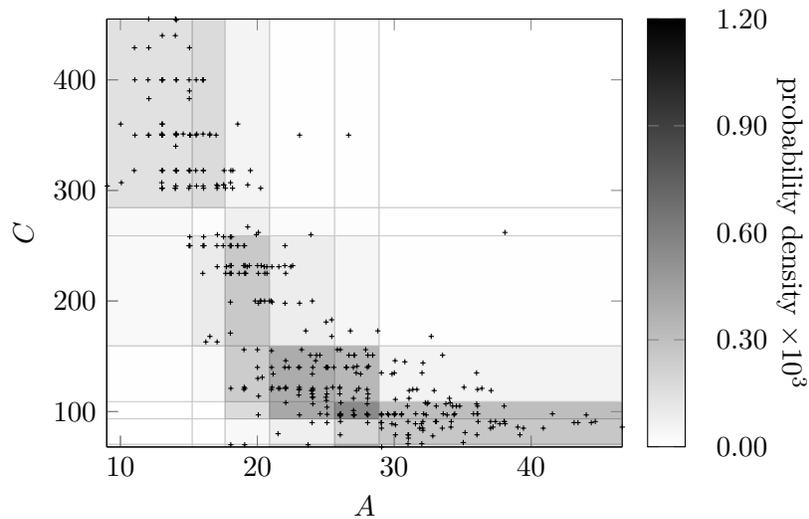}
  \caption{
    Comparison of the Bayesian discretization policy for variables $A$ and $C$ to the original Auto MPG data learned with a fixed network.
    The marginal probability density closely matches the data.
  }
  \label{fig:auto_exp1_distr_1_3}
\end{figure}

Figure~\ref{fig:auto_exp1_distr_1_3} shows the marginal probability density for variables $A$ and $C$ under the Bayesian discretization policy overlaid with the original Auto MPG data.
The resulting probability density is a good match to the original data.

\subsubsection{Discretization while Learning Structure}
\label{subsubsec:auto_exp2}

\begin{figure}[ht]
  \centering
  \scalebox{0.9}{
%
%
%
%


\begin{tikzpicture}[
		>={Stealth[round]}
      ]

      \node[obs]                 (c2) {$B_{{(5)}}$};
      \node[latent, below left = 2.0 and 1.0 of c2]   (c3){$C_{{(6)}}$};
      \node[latent, left =5.0 of c2]   (c5){$E_{{(4)}}$};
      \node[latent, below left = 1.75 and 1.0 of c3] (c4){$D_{{(2)}}$};
      \node[obs, below left = 1.0 and 2.0 of c2] (c8){$H_{{(3)}}$};
      \node[latent, below = 0.5 of c4] (c6){$F_{{(2)}}$};
      \node[latent, left = 1.5 of c8] (c1){$A_{{(3)}}$};
      \node[obs, below left = 1.0  and 1.0 of c8] (c7) {$G_{{(13)}}$};

  \edge {c2}{c3,c5,c1,c8,c3};
  \draw[->] (c2) to[bend left](c4);
  \draw[->] (c2) to[bend left](c6);
  \edge {c5}{c1,c8};
  \draw[->] (c5) to[bend right](c4);
  \draw[->] (c5) to[bend right](c6);
  \edge {c8}{c7,c4,c3};
  \edge {c1}{c7};
  \edge {c3}{c4};
  \edge {c4}{c6};

\end{tikzpicture}

}
  \caption{The discrete-valued Bayesian network learned from the Auto MPG dataset using the Bayesian method.}
  \label{fig:auto_graph_2}
\end{figure}
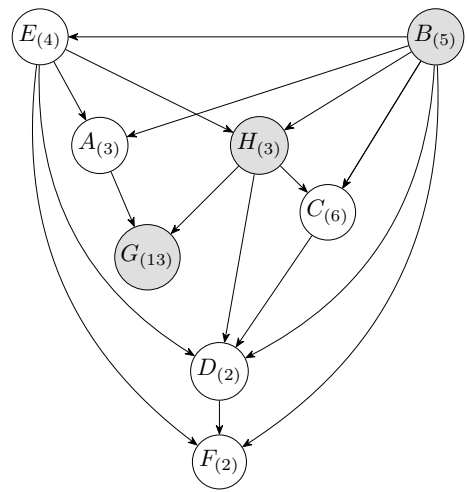

In this experiment, the network structure was not fixed in advance and was learned simultaneously with the discretization policies.
Figure~\ref{fig:auto_graph_2} shows a learned Bayesian network structure and the corresponding numbers of intervals after discretization for each continuous variable.
This result was obtained by running Algorithm~\ref{alg:structure_learn} fifty times using the Bayesian method and choosing the structure with the highest K2 score (Equation \ref{eq:BayesianScore}).

\begin{figure}[ht]
  \centering
  \input{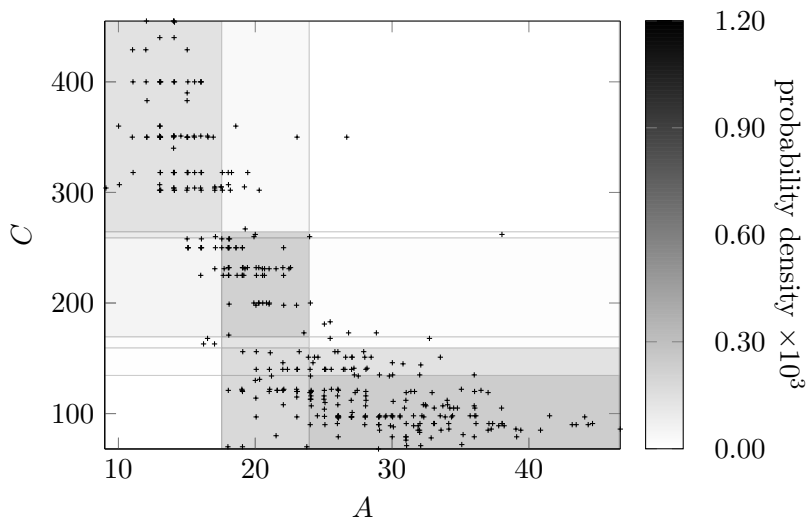}
  \caption{
    Comparison of the Bayesian discretization policy for variables $A$ and $C$ to the original Auto MPG data learned simultaneously with the network structure.
    Although the number of discretization edges is less than those in Figure~\ref{fig:auto_exp1_distr_1_3}, the probability distribution still closely matches the original data.
  }
  \label{fig:auto_exp2_distr_1_3}
\end{figure}

Figure~\ref{fig:auto_exp2_distr_1_3} compares the Bayesian discretizaton policy for variables $A$ and $C$ in the learned network with the original Auto MPG data.
The color of a discretized region indicates the marginal probability density of a sample from $P(A,C)$ being drawn from that region.
Although there are fewer discretization edges for $A$ and $C$ in the learned network, the marginal distribution is still captured.
The discretization policy will vary as the network structure changes, and it still produces high-quality discretizations.

%
\subsection{Dataset 2: Wine}
\label{subsec:wine}

The Wine dataset contains variables related to the chemical analysis of wines from three different Italian cultivars.
The dataset has \num{178} samples over fourteen variables.
Variable $A$ is the only discrete variable and has three instantiations.

\subsubsection{Discretization with Fixed Structure}
\label{subsubsec:wine_exp1}

\begin{figure}[ht]
  \centering
  \scalebox{0.8}{
%
%
%
%


\begin{tikzpicture}[
      >={Stealth[round]}
      ]

      \node[latent]                 (c8) {$H$};
      \node[obs, below left=0.75 and 0.0 of c8]   (c1){$A_{{(3)}}$};
      \node[latent, below left=0.75 and 4.0 of c8]   (c7){$G$};
      \node[latent, below right=0.75 and 0.4 of c1] (c13){$M$};
      \node[latent, below right=0.75 and 3.5 of c1] (c14){$N$};
      \node[latent, below left=0.75 and 2.0 of c1] (c12){$L$};
      \node[latent, below left=0.75 and 3.5 of c1] (c11){$K$};
      \node[latent, below left=0.75 and 5.5 of c1] (c4){$D$};
      \node[latent, below left=0.0 and 6.0 of c8] (c9){$I$};
      \node[latent, below left=0.75 and 0.5 of c12] (c5){$E$};
      \node[latent, below left=0.75 and 0.0 of c14] (c6){$F$};
      \node[latent, below right=0.0 and 1.5 of c8] (c10){$J$};
      \node[latent, below right=0.0 and 1.0 of c5] (c2){$B$};   
		\node[latent, below right=0.0 and 2.0 of c5] (c3){$C$};         
      

 \edge {c8}{c1,c7,c9,c10,c13};
 \edge {c1}{c4,c11,c12,c2,c3,c14,c13,c9,c10};
 \edge {c1,c4}{c5};
 \edge {c4}{c9};
 \edge {c1,c14}{c6};
 \edge {c6}{c10};

\end{tikzpicture}

}
  \caption{Bayesian network structure obtained from running K2 on the initially uniformly discretized Wine dataset.}
  \label{fig:wine_graph_1}
\end{figure}
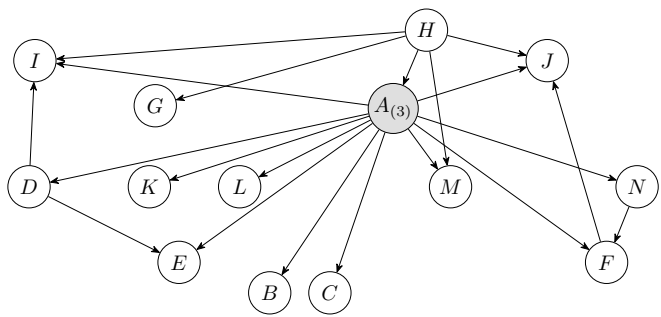

The Bayesian and MDL discretization methods were tested on the Wine data using the network shown in Figure~\ref{fig:wine_graph_1}.
This structure was obtained by initially discretizing each continuous variable into three uniform-width intervals, where three is the median cardinality of the discrete variables, and then taking the structure with highest likelihood from \num{1000} runs of K2.

Table~\ref{table:wine_disc_table_1} lists the discretization edges and mean log-likelihoods under \num{10}-fold cross validation of the Bayesian network resulting from each discretization method.
The Bayesian method outperforms the MDL method in likelihood by a significant margin.
The MDL method creates significantly fewer discretization edges.

Some discretization edges appear in both discretization methods, such as \num{1.42} and \num{2.35} for variable $C$ and \num{0.785} for variable $L$. MDL can find some important discretization edges, but it is not sensitive to find more edges. Actually, as shown later in Section \ref{subsec:discuss_exp}, while the MDL and Bayesian methods both produce the optimal discretization as the training data grows, the Bayesian method can produce these discretizations with fewer data samples.

\begin{table}
  \centering
  \caption{
    Results from discretization of the Wine MPG dataset with fixed structure.
    Bold discretization edges were identified by both methods.
  }
  \scalebox{1.0}{
  \scriptsize
\sisetup{detect-weight = true}
\begin{tabular}{@{}ccc@{}}
\toprule
Variable & Bayesian            & MDL          \\
\midrule
$B$    & \num{12.745}, \num{13.54}                         & \num{12.78}        \\
$C$    & \textbf{\num{1.42}}, \textbf{\num{2.235}}                           & \textbf{\num{1.42}}, \textbf{\num{2.235}}  \\
$D$    & \num{2.03}, \num{2.605}, \num{3.07}               & -            \\
$E$    & \textbf{\num{17.9}}, \num{23.25}                           & \textbf{\num{17.9}}         \\
$F$    & \num{88.5}, \num{135.0}                           & -            \\
$G$    & \num{2.105}, \num{2.58}, \num{3.01}               & -            \\
$H$    & \num{0.975}, \num{1.885}, \num{2.31}, \num{3.355} & -            \\
$I$    & \textbf{\num{0.395}}                                       & \textbf{\num{0.395}}       \\
$J$    & \num{1.185}, \num{1.655}                          & -            \\
$K$    & \textbf{\num{3.46}}, \num{4.85}, \num{7.4}                 & \textbf{\num{3.46}}, \num{7.55}   \\
$L$    & \textbf{\num{0.785}}, \num{1.005}, 1.295                   & \textbf{\num{0.785}}        \\
$M$    & \num{2.475}                                       & \num{2.115}, \num{2.505} \\
$N$    & \num{476.0}, \num{716.0}, \num{900.5}             & -            \\
\midrule
Log-Likelihood   & \num{-19.94} & \num{-23.60}        \\
\bottomrule
\end{tabular}
  }
  \label{table:wine_disc_table_1}
\end{table}

\begin{figure}[ht]
  \centering
  \input{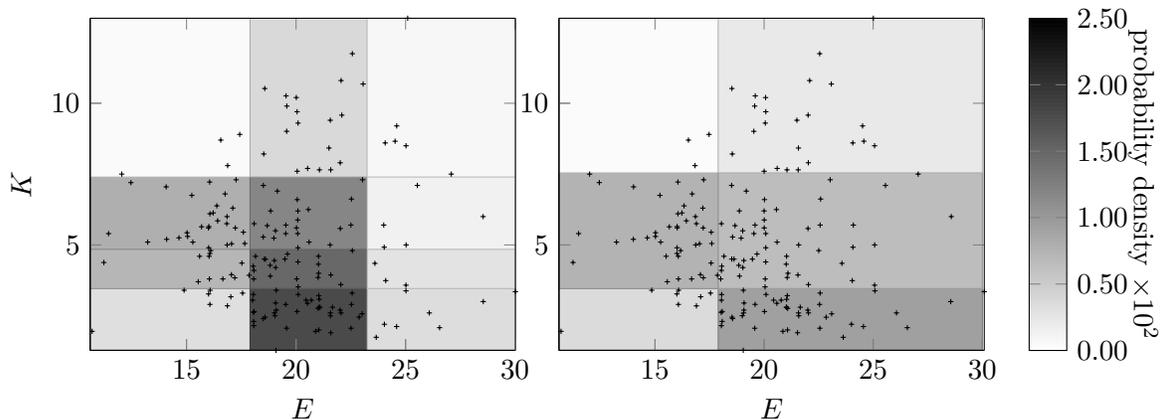}
  \caption{Comparison of the discretization policies for variables $E$ and $K$ obtained using the Bayesian and MDL methods.}
  \label{fig:wine_exp1_distr}
\end{figure}

Figure~\ref{fig:wine_exp1_distr} compares the Bayesian and MDL discretizaton policies for variables $E$ and $K$ with the original Wine data.
The discretization edges \num{17.9} for $E$ and \num{34.6} for $K$ appear in both plots.
The MDL method does not use enough intervals for discretization.
Relative sensitivities of each method to the input data is discussed in Section~\ref{subsec:discuss_exp}.

\subsubsection{Discretization while Learning Structure}
\label{subsubsec:wine_exp2}

\begin{figure}[ht]
  \centering
  \scalebox{0.7}{
%
%
%
%


\begin{tikzpicture}[
      >={Stealth[round]}
      ]
      \node[latent]  at  (0,0)           (c2){$B_{{(2)}}$};
	  \node[latent]  at  (-3,5.25)      (c3) {$C_{{(2)}}$};
	  \node[latent]  at  (1.5,5.24)        (c4){$D_{{(3)}}$};
	  \node[latent]  at   (3,3.75)       (c10){$J_{{(5)}}$};
	  \node[latent]  at   (0,3)          (c9){$I_{{(1)}}$};
	  \node[latent]  at   (0,1.5)          (c14){$N_{{(3)}}$};
	  \node[latent]  at   (3,0)        (c7){$G_{{(2)}}$};	  
	  \node[latent]  at   (2.5,-3)        (c5){$E_{{(2)}}$};
	  \node[latent]  at   (-1.5,-3)      (c8){$H_{{(5)}}$};
	  \node[latent]  at   (-3,3.75)     (c12){$L_{{(2)}}$};
	  \node[latent]  at   (-3,1)        (c13){$M_{{(3)}}$};
	  \node[obs]     at    (-3,-1.5)      (c1){$A_{{(3)}}$};
	  \node[latent]  at    (-3,-3)      (c11){$K_{{(4)}}$};
	  \node[latent]  at    (-5.5,-3)      (c6){$F_{{(2)}}$};


  \edge {c3}{c12,c10,c9,c2};
  \edge {c4}{c10,c7,c9,c5}
  \edge {c10}{c9,c7,c13};
  \edge{c9}{c14,c6,c5};
  \draw[->] (c9) to[bend right](c2);
  \edge {c14}{c2,c7};
  \edge {c7}{c2,c8};
  \edge {c12}{c13,c14};
  \edge {c13}{c1,c5};
  \edge {c1}{c6,c11,c8,c5};
  \edge {c2}{c1};
  \draw[->] (c12) to[bend right](c1);

\end{tikzpicture}

}
  \caption{The discrete-valued Bayesian network learned from the Wine dataset using the Bayesian method.}
  \label{fig:wine_graph_2}
\end{figure}
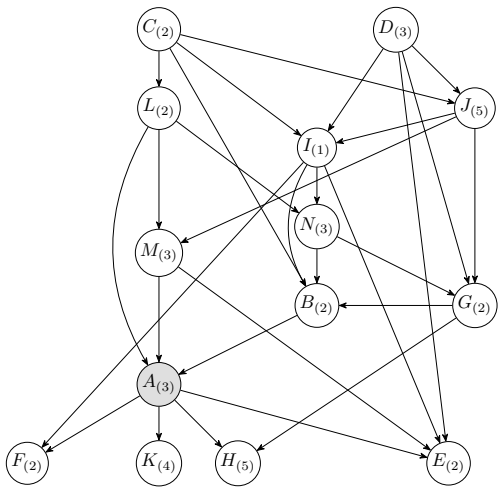

Figure~\ref{fig:wine_graph_2} is the discrete-valued Bayesian network learned from the Wine dataset, obtained by running Algorithm~\ref{alg:structure_learn} fifty times.
A comparison of Figures~\ref{fig:wine_graph_2} and \ref{fig:wine_graph_1} show that the Bayesian network learned during the discretization process has more edges than the network learned on the initially discretized data.
When a network is learned along with discretization, the algorithm has more freedom to adjust the structure and the discretization policy simultaneously to identify useful correlations and produce a denser structure.

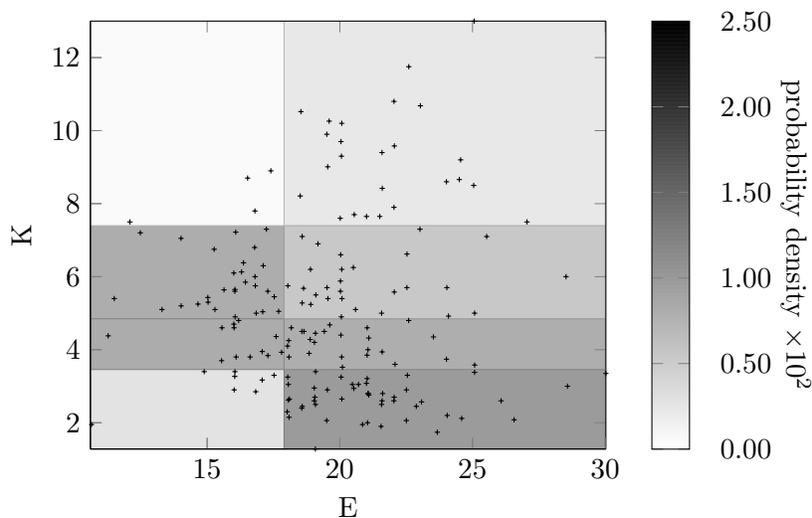
\begin{figure}[ht]
  \centering
  \begin{tikzpicture}
		\begin{axis}[
			view={0}{90},
			enlargelimits=0,
			zmin=0.0, zmax=1.0,
			xlabel=E,
			ylabel=K,
			colorbar,
			point meta min=0, point meta max=2.5,
			colormap={}{ gray(0cm)=(1); gray(1cm)=(0);},
			colorbar style={
			        ytick={0,0.5,...,2.51},
			        ylabel={probability density $\times 10^2$},
			        y label style={at={(axis description cs:4.5,0.5)},rotate=180,anchor=north},
			        yticklabel style={
			            text width=2.5em,
			            align=right,
			            /pgf/number format/.cd,
			                fixed,
			                fixed zerofill
			        }
			    }
			]

			\addplot3[patch,shader=faceted,patch type=rectangle] coordinates {(10.6000,1.2800,0.276486) (17.9000,1.2800,0.276486) (17.9000,3.4600,0.2765) (10.6000,3.4600,0.2765)};
			\addplot3[patch,shader=faceted,patch type=rectangle] coordinates {(10.6000,3.4600,0.808121) (17.9000,3.4600,0.808121) (17.9000,4.8500,0.8081) (10.6000,4.8500,0.8081)};
			\addplot3[patch,shader=faceted,patch type=rectangle] coordinates {(10.6000,4.8500,0.795058) (17.9000,4.8500,0.795058) (17.9000,7.4000,0.7951) (10.6000,7.4000,0.7951)};
			\addplot3[patch,shader=faceted,patch type=rectangle] coordinates {(10.6000,7.4000,0.047701) (17.9000,7.4000,0.047701) (17.9000,13.0000,0.0477) (10.6000,13.0000,0.0477)};

			\addplot3[patch,shader=faceted,patch type=rectangle] coordinates {(17.9000,1.2800,0.972401) (30.0000,1.2800,0.972401) (30.0000,3.4600,0.9724) (17.9000,3.4600,0.9724)};
			\addplot3[patch,shader=faceted,patch type=rectangle] coordinates {(17.9000,3.4600,0.784827) (30.0000,3.4600,0.784827) (30.0000,4.8500,0.7848) (17.9000,4.8500,0.7848)};
			\addplot3[patch,shader=faceted,patch type=rectangle] coordinates {(17.9000,4.8500,0.531518) (30.0000,4.8500,0.531518) (30.0000,7.4000,0.5315) (17.9000,7.4000,0.5315)};
			\addplot3[patch,shader=faceted,patch type=rectangle] coordinates {(17.9000,7.4000,0.227273) (30.0000,7.4000,0.227273) (30.0000,13.0000,0.2273) (17.9000,13.0000,0.2273)};

			\addplot3[mark=+, draw=none, mark size=1.0, jitter = 0.1] coordinates {
(15.600,5.640,1) (11.200,4.380,1) (18.600,5.680,1) (16.800,7.800,1) (21.000,4.320,1) (15.200,6.750,1) (14.600,5.250,1) (17.600,5.050,1) (14.000,5.200,1) (16.000,7.220,1) (18.000,5.750,1) (16.800,5.000,1) (16.000,5.600,1) (11.400,5.400,1) (12.000,7.500,1) (17.200,7.300,1) (20.000,6.200,1) (20.000,6.600,1) (16.500,8.700,1) (15.200,5.100,1) (16.000,5.650,1) (18.600,4.500,1) (16.600,3.800,1) (17.800,3.930,1) (20.000,3.520,1) (25.000,3.580,1) (16.100,4.800,1) (17.000,3.950,1) (19.400,4.500,1) (16.000,4.700,1) (22.500,5.700,1) (19.100,6.900,1) (17.200,3.840,1) (19.500,5.400,1) (19.000,4.200,1) (20.500,5.100,1) (15.500,4.600,1) (18.000,4.250,1) (15.500,3.700,1) (13.200,5.100,1) (16.200,6.130,1) (18.800,4.280,1) (15.000,5.430,1) (17.500,4.360,1) (17.000,5.040,1) (18.900,5.240,1) (16.000,4.900,1) (16.000,6.100,1) (18.800,6.200,1) (17.400,8.900,1) (12.400,7.200,1) (17.200,5.600,1) (14.000,7.050,1) (17.100,6.300,1) (16.400,5.850,1) (20.500,6.250,1) (16.300,6.380,1) (16.800,6.000,1) (16.700,6.800,1) (10.600,1.950,1) (16.000,3.270,1) (16.800,5.750,1) (18.000,3.800,1) (19.000,4.450,1) (19.000,2.950,1) (18.100,4.600,1) (15.000,5.300,1) (19.600,4.680,1) (17.000,3.170,1) (16.800,2.850,1) (20.400,3.050,1) (25.000,3.380,1) (24.000,3.740,1) (30.000,3.350,1) (21.000,3.210,1) (16.000,3.800,1) (16.000,4.600,1) (18.000,2.650,1) (14.800,3.400,1) (23.000,2.570,1) (19.000,2.500,1) (18.800,3.900,1) (24.000,2.200,1) (22.500,4.800,1) (18.000,3.050,1) (18.000,2.620,1) (22.800,2.450,1) (26.000,2.600,1) (21.600,2.800,1) (23.600,1.740,1) (18.500,2.400,1) (22.000,3.600,1) (20.700,3.050,1) (18.000,2.150,1) (18.000,3.250,1) (19.000,2.600,1) (21.500,2.500,1) (16.000,2.900,1) (18.500,4.500,1) (18.000,2.300,1) (17.500,3.300,1) (18.500,2.450,1) (21.000,2.800,1) (19.500,2.060,1) (20.500,2.940,1) (22.000,2.700,1) (19.000,3.400,1) (22.500,3.300,1) (19.000,2.700,1) (20.000,2.650,1) (19.500,2.900,1) (21.000,2.000,1) (20.000,3.800,1) (21.000,3.080,1) (22.500,2.900,1) (21.500,1.900,1) (20.800,1.950,1) (22.500,2.060,1) (16.000,3.400,1) (19.000,1.280,1) (20.000,3.250,1) (28.500,6.000,1) (26.500,2.080,1) (21.500,2.600,1) (21.000,2.800,1) (21.000,2.760,1) (21.500,3.940,1) (28.500,3.000,1) (24.500,2.120,1) (22.000,2.600,1) (18.000,4.100,1) (20.000,5.400,1) (24.000,5.700,1) (21.500,5.000,1) (17.500,5.450,1) (18.500,7.100,1) (21.000,3.850,1) (25.000,5.000,1) (19.500,5.700,1) (24.000,4.920,1) (21.000,4.600,1) (20.000,5.600,1) (23.500,4.350,1) (20.000,4.400,1) (18.500,8.210,1) (21.000,4.000,1) (20.000,4.900,1) (21.500,7.650,1) (21.500,8.420,1) (21.500,9.400,1) (24.000,8.600,1) (22.000,10.800,1) (25.500,7.100,1) (18.500,10.520,1) (20.000,7.600,1) (22.000,7.900,1) (19.500,9.010,1) (27.000,7.500,1) (25.000,13.000,1) (22.500,11.750,1) (21.000,7.650,1) (20.000,5.880,1) (22.000,5.580,1) (18.500,5.280,1) (22.000,9.580,1) (22.500,6.620,1) (23.000,10.680,1) (19.500,10.260,1) (24.500,8.660,1) (25.000,8.500,1) (19.000,5.500,1) (19.500,9.900,1) (20.000,9.700,1) (20.500,7.700,1) (23.000,7.300,1) (20.000,10.200,1) (20.000,9.300,1) (24.500,9.200,1) };
		\end{axis}
	\end{tikzpicture}
  \caption{
    The discretization policy for variables $E$ and $K$ on the learned network.
    The discretization edge $E = 17.9$ also appears in the policies for the fixed network.
  }
  \label{fig:wine_exp2_distr}
\end{figure}

Figure~\ref{fig:wine_exp2_distr} shows the discretization policy for variables $E$ and $K$ obtained with the Bayesian method.
The discretization edge $E = 17.9$ also appears in both discretization policies from the fixed network (Figure~\ref{fig:wine_exp1_distr}).
This suggests that discretization edges can be robust against network structure.
Furthermore, the discretization edge at $E = 23.5$ in the fixed-structure case is missing in the learned structure.
This is caused by $E$ having twice as many parents in the learned network structure.
The more parents a variable has, the fewer discretization intervals it can to support, as the number of sufficient statistics required to define the resulting distribution increases exponentially with the number of parents.

%
%
\subsection{Dataset 3: Housing}
\label{subsec:housing}

The Housing dataset contains variables related to the values of houses in Boston suburbs.
The dataset has \num{506} samples over fourteen variables.
Only variables $D$ and $I$ are discrete-valued, with \num{2} and \num{9} instantiations, respectively.
Despite their being continuous, several variables in the Housing dataset possess many repeated values.
The following experiments were conducted with a maximum of three parents per variable to prevent running out of memory when running MDL.

\subsubsection{Discretization with Fixed Structure}
\label{subsubsec:housing_exp1}

\begin{figure}[ht]
  \centering
  \scalebox{0.8}{
%
%
%
%


\begin{tikzpicture}[
      >={Stealth[round]}
      ]

	  \node[latent]          (c5) {$E$};
	  \node[obs, right = 3.0 of c5]  (c4){$D_{{(2)}}$};
	  \node[latent, below left = 1.0 and 1.0 of c5]  (c2){$B$};
	  \node[obs, below = 2.0 of c5]  (c9){$I_{{(9)}}$};
	  \node[latent, below = 1.0 of c9]  (c10){$J$};
	  \node[latent, below = 1.0 of c10] (c12){$L$};
	  \node[latent, below right = 1.0 and 2.0 of c10]  (c11){$K$};
	  \node[latent, below right = 2.0 and 2.0 of c5] (c3){$C$};
	  \node[latent, below right = 1.0 and 3.0 of c5] (c8){$H$};
	  \node[latent, left = 1.5 of c2] (c7){$G$};
	  \node[latent, below = 1.0 of c7] (c13){$M$};
	  \node[latent, below = 1.0 of c13] (c14){$N$};
	  \node[latent, right = 1.5 of c14] (c1){$A$};
	  \node[latent, below right = 0.75 and 0.5 of c14] (c6){$F$};


  \edge {c5}{c2,c3,c8,c9};
  \draw[->] (c5) to[bend right](c7);
  \edge {c2}{c9,c3,c7}
  \draw[->] (c2) to[bend right](c10);
  \edge {c3}{c9,c10,c11};
  \edge{c9}{c10,c11};
  \edge {c10}{c11,c12};
  \edge {c7}{c13};
  \edge {c13}{c14};
  \edge {c14,c1}{c6};

\end{tikzpicture}

}
  \caption{Bayesian network structure obtained from running K2 on the initially uniformly discretized Housing dataset.}
  \label{fig:housing_graph_1}
\end{figure}
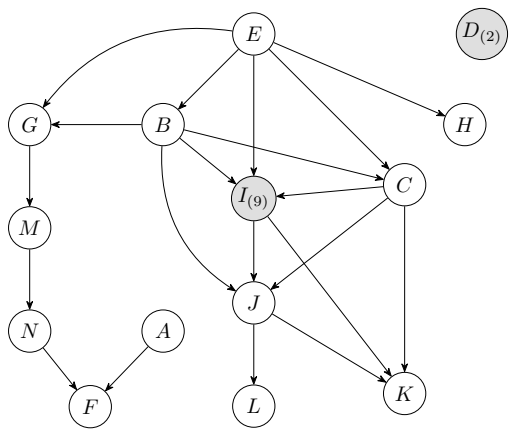

The Bayesian network structure in Figure~\ref{fig:housing_graph_1} was obtained by initially discretizing each continuous variable into five uniform-width intervals and then running the K2 algorithm \num{1000} times and choosing the network with the highest likelihood.
Table~\ref{table:housing_disc_table_1} shows the numbers of interval after discretization for each continuous variable and the log-likelihood of the dataset based on each discretization method.
The MDL method does not produce any discretization edges for most variables.
The relative weighting of each method's objective function will be discussed in Section~\ref{subsec:discuss_exp}.


\begin{table}
  \centering
  \caption{
    Discretization policy summary of the Housing dataset based on the fixed network structure shown.
    The first twelve rows show the numbers of discretization intervals and the last row is the mean cross-validated log-likelihood.
  }
  \scriptsize
\begin{tabular}{@{}ccc@{}}
\toprule
Variable   & Bayesian & MDL          \\
\midrule
$A$    & 3        & 1      \\
$B$    & 4        & 1      \\
$C$    & 8        & 1      \\
$E$    & 14       & 1      \\
$F$    & 4        & 1      \\
$G$    & 3        & 1      \\
$H$    & 6        & 1      \\
$J$    & 8        & 1      \\
$K$    & 5        & 7      \\
$L$    & 4        & 1      \\
$M$    & 6        & 1      \\
$N$    & 6        & 1      \\
\midrule
Log-Likelihood   & $-31.40$   & $-43.20$ \\
\bottomrule
\end{tabular}

  \label{table:housing_disc_table_1}
\end{table}

\begin{figure}[ht]
  \centering
  \input{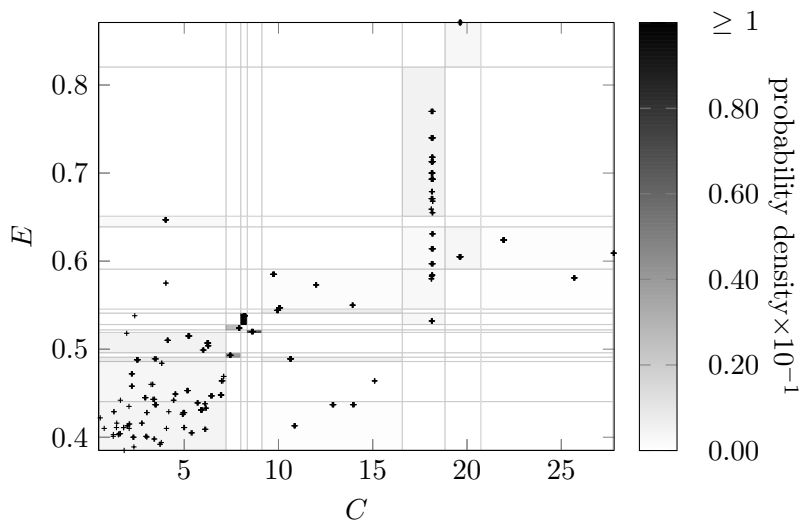}
  \caption{
    The discretization policies for variables $C$ and $E$ learned using the Bayesian approach on the fixed network.
    The scatter points were jittered to reveal the repeated values in the dataset.
    Values perfectly repeated in $C$ and $E$ lead to high correlation and concentrated densities in the marginal distribution.
  }
  \label{fig:housing_exp1_distr_3_5}
\end{figure}

Figures~\ref{fig:housing_exp1_distr_3_5} and \ref{fig:housing_exp1_distr_8_5} show the discretization policy learned using the Bayesian approach on the fixed network shown in Figure~\ref{fig:housing_graph_1}.
The scatter points in Figure~\ref{fig:housing_exp1_distr_3_5} were jittered to show the quantity of repeated values for variables $C$ and $E$.
Each repeated point forms a single discrete region, thereby encouraging discretization.
In contrast, the samples for $H$ are well spread out, resulting in fewer discretization regions and larger discretization intervals.

\subsubsection{Discretization while Learning Structure}
\label{subsubsec:housing_exp2}

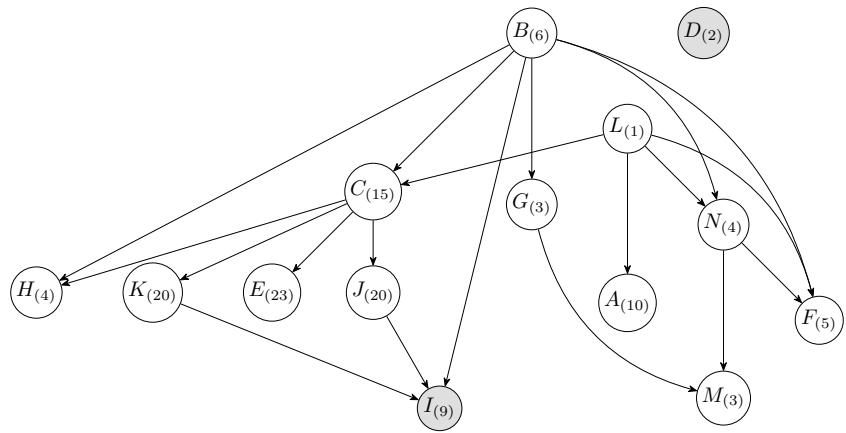
\begin{figure}[H]
  \centering
  \scalebox{0.8}{
%
%
%
%


\begin{tikzpicture}[
      >={Stealth[round]}
      ]

	  \node[latent]          (c2) {$B_{(6)}$};
	  \node[obs, right = 2.0 of c2]  (c4){$D_{(2)}$};
	  \node[latent, below right = 1.0 and 1.0 of c2]  (c12){$L_{(1)}$};
	  \node[latent, below right = 1.0 and 1.0 of c12]  (c14){$N_{(4)}$};
	  \node[latent, below right = 1.0 and 1.0 of c14]  (c6){$F_{(5)}$};
	  \node[latent, below= 2.0 of c2]  (c7){$G_{(3)}$};
	  \node[latent, below = 2.0 of c12]  (c1){$A_{(10)}$};
	  \node[latent, below = 2.0 of c14]  (c13){$M_{(3)}$};
	  \node[latent, below left = 2.0 and 2.0 of c2] (c3){$C_{(15)}$};
	  \node[latent, below = 0.75 of c3]  (c10){$J_{(20)}$};
	  \node[latent, below left = 1.0 and 1.0 of c3] (c5){$E_{(23)}$};
	  \node[latent, left = 1.0 of c5] (c11){$K_{(20)}$};
	  \node[latent, left = 1.0 of c11] (c8){$H_{(4)}$};
	  \node[obs, below right = 3 and 0.5 of c3] (c9){$I_{(9)}$};


  \edge {c2}{c9,c8,c3,c7};
  \draw[->] (c2) to[bend left](c14);
  \draw[->] (c2) to[bend left](c6);
  \edge {c12}{c3,c1,c14};
  \draw[->] (c12) to[bend left](c6);
  \edge {c14}{c13,c6};
  \draw[->] (c7) to[bend right](c13);
  \edge {c3} {c8,c11,c5,c10};
  \edge {c11,c10}{c9};

\end{tikzpicture}

}
  \caption{
    The discrete-valued Bayesian network learned from the Housing dataset using the optimal Bayesian method.
  }
  \label{fig:housing_graph_2}
\end{figure}

Figure~\ref{fig:housing_graph_2} shows a learned Bayesian network structure and the corresponding numbers of intervals after discretization for each continuous variable.
Variable $D$ is neighborless in both the learned and fixed networks.
The continuous variables $C$, $E$, $J$ and $K$, which are all connected through $C$, have many discretization intervals.
Typically, when discretizing a variable, the expected number of discretization intervals is close to the highest cardinality among variables in its Markov blanket.
This naturally leads to clusters of variables with many discretization intervals.

\begin{figure}[H]
  \centering
  \input{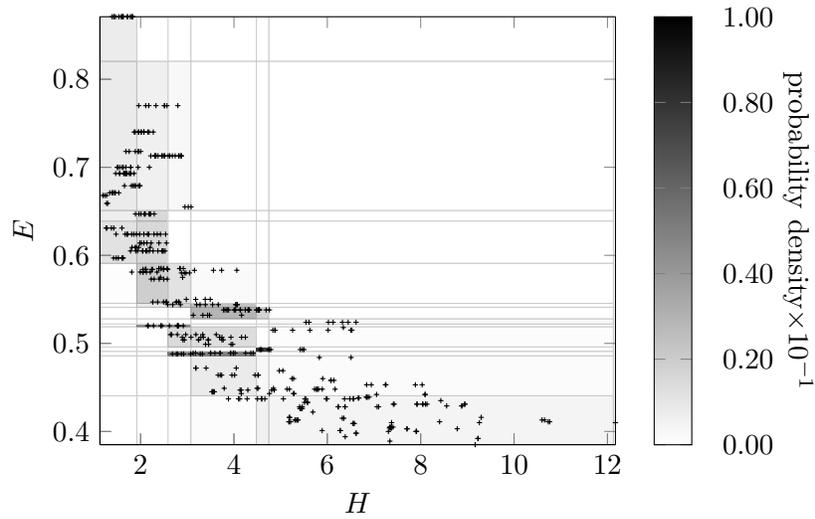}
  \caption{
    The discretization policies for variables $H$ and $E$ learned using the optimal Bayesian approach on the fixed network shown in Figure~\ref{fig:housing_graph_1}.
    The values for variable $H$ are not as repeated, thus producing fewer discretization intervals than $E$.
  }
  \label{fig:housing_exp1_distr_8_5}
\end{figure}

\begin{figure}[H]
  \centering
  \input{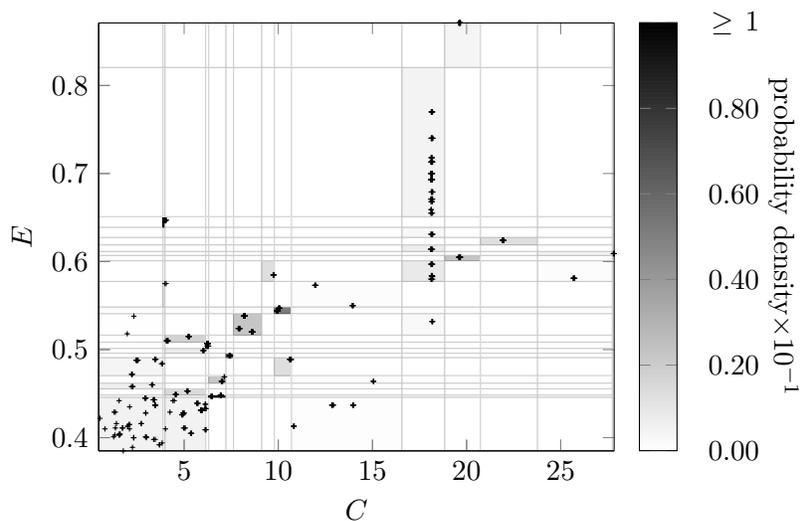}
  \caption{
    The discretization policy for variables $C$ and $E$ on the learned network in Figure~\ref{fig:housing_graph_2}.
    The large number of discretization edges are due to repeated values in $C$ and $E$.
    The two regions at $(C,E) = (3.9,0.65)$ and $(6.8,0.44)$ are color-saturated due to the high concentration of repeated values.
  }
  \label{fig:housing_exp2_distr_3_5}
\end{figure}

\begin{figure}[H]
  \centering
  \input{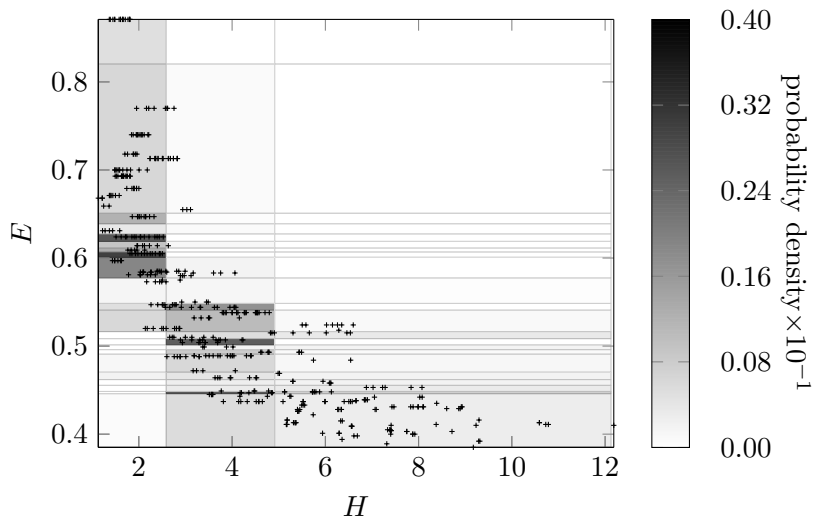}
  \caption{
    The discretization policy for variables $H$ and $E$ on the learned network in Figure~\ref{fig:housing_graph_2}.
    There are fewer discretization intervals for $H$ than for the fixed network structure in Figure~\ref{fig:housing_exp1_distr_8_5}, but the discretization result still closely models the raw data.
  }
  \label{fig:housing_exp2_distr_8_5}
\end{figure}

Figures~\ref{fig:housing_exp2_distr_3_5} and \ref{fig:housing_exp2_distr_8_5} show the discretization result for variables in the network shown in Figure~\ref{fig:housing_graph_2}.
Again, variable $C$ and $E$ have many discretization edges due to repeated values.
Although the number of intervals after discretization on $H$ is less the number in Figure~\ref{fig:housing_exp1_distr_8_5}, it still captures the distribution of the raw data along with the discretization edges for $E$.

\subsection{Results on Additional Datasets}

Six additional datasets \citep{Lichman_2013} are tested to compare the Bayesian and MDL discretization methods.
The results are summarized in Table~\ref{table:more_data_set}.
Each row corresponds to a dataset and provides the number of samples $n$, the number of variables $N$, the fixed network structure, and the log-likelihood for each method.
Fixed network structures are obtained by the same principle as with the previous three datasets and are represented by listing the parents for each variable.
For example, in the first row, the representation $\curly{\emptyset,A,BE,AB,B,AH,BDG,B}$ indicates that $\Pa_A = \emptyset$, $\Pa_B = A$, $\Pa_c = \curly{B,E}$, etc.
The Bayesian discretization method has higher mean cross-validated log likelihood in all cases.

\begin{table}[ht]
	\centering
	\caption{
		Discretization results for several datasets, summarizing the number of samples, number of variables, network structure, and mean cross-validated log-likelihood under each discretization method.
    The Bayesian method consistently has higher likelihood.
	}
  \resizebox{\textwidth}{!}{
	 \scriptsize
\begin{tabular}{lrrcSS}
	\toprule
	Name                    & \multicolumn{1}{c}{$n$} & \multicolumn{1}{c}{$N$} & Network & {Bayesian}&  {MDL} \\
	\midrule
	Seeds                   & 210  & 8   &  \scalebox{1.0}{$\curly{\emptyset,A,BE,AB,B,AH,BDG,B}$} & -2.35            & -5.71       \\
	Glasses                 & 214  & 10  & \scalebox{1.0}{$\curly{F,EFH,FGH,AFH,AF,\emptyset,AF,F,\emptyset,\emptyset}$}        & -1.62            & -7.75       \\
	Heart                   & 270  & 13  & \scalebox{1.0}{$\curly{H,M,J,EF,\emptyset,\emptyset,J,IK,C,\emptyset,IJ,A,I}$}  & -25.81           & -26.91      \\
	Liver Disorder          & 345  & 7   & \scalebox{1.0}{$\curly{\emptyset,EF,\emptyset,CE,C,CD,D}$}   & -23.73           & -26.57      \\
	Indian Liver Patient    & 583  & 10  &  \scalebox{1.0}{$\curly{DEFI,E,\emptyset,\emptyset,DG,\emptyset,F,CGIJ,DEJ,\emptyset,\emptyset}$}  & -34.89           & -40.53      \\
	Banknote Authority      & 1372 & 5   &  \scalebox{1.0}{$\curly{BD,D,AB,\emptyset,AD}$}  & -8.99           & -12.13  \\
	\bottomrule
\end{tabular}
  }
	\label{table:more_data_set}
\end{table}

Finding optimal discretization policies according to Equation~\ref{eq:opt_prob} via dynamic programming requires computation times on the order of days on a personal laptop on datasets with more than \num{100} variables.
The runtime for the MDL approach is significantly worse.
Approximation approaches are reasonable options for such large datasets, for example a greedy heuristic~\citep{Friedman_1996}.
Approximations can also be applied during dynamic programming when solving Equation~\ref{eq:opt_prob} \citep{Boulle_2006}.

\subsection{Discussion}
\label{subsec:discuss_exp}

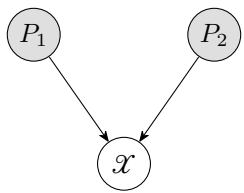
\begin{figure}[ht]
  \centering
%
%
%
%


\begin{tikzpicture}[
      >={Stealth[round]}
      ]

  \node[latent]                               (y) {$\cX$};
  \node[obs, above=of y, xshift=-1.2cm] (p1) {$P_1$};
  \node[obs, above=of y, xshift=1.2cm]  (p2) {$P_2$};

  \edge {p1,p2} {y} ;


\end{tikzpicture}

  \caption{A simple Bayesian network used to demonstrate the sensitivity of each method.}
  \label{fig:exp_discuss}
\end{figure}

To qualitatively assess the sensitivity of each method to the number and position of discretization intervals, consider the Bayesian network in Figure~\ref{fig:exp_discuss}, where $\cX$ is continuous and $P_1$ and $P_2$ are discrete.
If $\Pa_{\cX} = \curly{P_1, P_2}$ and $D_{\cX} = \curly{1,2,3, \ldots n}$, then the corresponding objective functions for the discretization methods are:
\begin{small}
  \begin{equation}
  \begin{aligned}
  f_{\text{MDL}} & = \overbrace{Z_1 \cdot k + \ln(k) + \ln {{n +k - 1}\choose{k -1}} }^{\text{Penalty Term}}+   \overbrace{\rule{0pt}{1.35em} n \cdot I(X^*, \Pa_X)}^{\text{Edge Position Term}} \\
  f_{\text{Bayesian}} &= \underbrace{{\vrule width0pt height0pt depth1.48em\relax}  Z_2 \cdot k + \sum_{i=1}^k \ln {{\gamma_i + J_\text{P} - 1}\choose{J_\text{P} -1}}}_{\text{Penalty Term}} + \underbrace{\sum^k_{i=1} \ln \left(   {{\gamma_i !}\over{n^{\text{P}}_{i,1}! n^{\text{P}}_{i,2}! \cdots n^{\text{P}}_{i,J_\text{P}}!    }} \right) }_{\text{Edge Position Term}}
  \end{aligned}
  \end{equation}
\end{small}

Here, $Z_1$ and $Z_2$ are constant over discretizations, $k$ is the number of discretization intervals, and $I\paren{A, B} = \sum_{a,b} \hat{P}(a,b) \ln\frac{\hat{P}(a,b)}{\hat{P}(a)\hat{P}(b)}$ is the mutual information based on estimated probabilities.
Note that the third term of $f_\text{MDL}$ was approximated using $H(p)$ (see Equation~\ref{eqn:MDL}) in the original work by \citet{Friedman_1996}, but it is written here without that approximation.

The penalty terms tend to increase with the number of discretization intervals; the edge position terms vary with the position and number of the discretization edges.
A policy with a larger number of discretization edges is only optimal if the edge position term varies enough with respect to the penalty term in order to produce a local minimum of sufficiently low value.

The value of the edge position term for MDL is primarily determined by the mutual information, which varies less severely than the corresponding terms in the Bayesian method.
The MDL term uses empirical probability distributions based off of ratios of counts, whereas the Bayesian method uses factorial terms, and thus the MDL method varies less.
The MDL method is therefore less sensitive to the discretization edges.
This sensitivity gives rise to the relative performance of the two methods in the experiments conducted above.

To support this argument, we test a synthetic dataset with $800$ samples using the network structure in Figure~\ref{fig:exp_discuss}.
Data samples are generated by the distribution described in Table~\ref{table:distri_exp}: both $P_1$ and $P_2$ are fair coins.
The continuous variable $\cX$ is conditionally dependent on $P_1$ and $P_2$.
Note that the symbol $\sim U[a,b]$ in Table~\ref{table:distri_exp} represents a uniform distribution from $a$ to $b$.
The optimal discretization policy on $\cX$ has six edges, $\curly{2.0,3.0,4.0,4.5,5.0,6.0}$, since $\cX$ takes numeric values from $1.0$ to $7.5$ and there are seven intervals corresponding to different joint distributions of $P_1$ and $P_2$.

\begin{table}[ht]
	\centering
	\caption{Probability distributions for the variables in Figure~\ref{fig:exp_discuss}.
           $P_1$ and $P_2$ are discrete whereas $\cX$ is continuous.}
	\label{table:distri_exp}
	\begin{tabular}{cl}
		\toprule
		Variable               & Distribution                                   \\ \midrule
		\multirow{2}{*}{$P_1$}&$P(p^{(1)}_1) = 0.5$     \\
		                                       & $P(p^{(2)}_1) = 0.5$     \\ \midrule
		\multirow{2}{*}{$P_2$}&$P(p^{(1)}_2) = 0.5$     \\
		                                       &$P(p^{(2)}_2) = 0.5$     \\ \midrule
		\multirow{4}{*}{$\cX$} & $p(\cx \mid p^{(1)}_1,p^{(1)}_2) \sim U[2,6]$\\
		                                     & $p(\cx \mid p^{(1)}_1,p^{(2)}_2) \sim U[1,3]$ \\
		                                     & $p(\cx \mid p^{(2)}_1,p^{(1)}_2) \sim U[4,5]$ \\
		                                     & $p(\cx \mid p^{(2)}_1,p^{(2)}_2) \sim U[4.5,7.5]$\\
		\bottomrule
	\end{tabular}
\end{table}

\begin{figure}
	\begin{subfigure}{0.45\textwidth}
		\centering
		\scalebox{0.6}{\begin{tikzpicture}
\begin{axis}[
		xlabel = Number of Samples,
		ylabel = Log-likelihood,
		legend style={at={(0.99,0.02)},anchor=south east},
		ymin= -1.2, ymax= -0.8
]
\addplot+[solid ,black,thick, mark options={black}] coordinates {
(150,	-1.03842)
(200,	-1.00487)
(250,	-0.978629)
(300,	-0.949342)
(350,	-0.931778)
(400,	-0.907931)
(450,	-0.886758)
(500,	-0.880402)
(550,	-0.859904)
(600,	-0.859312)
(650,	-0.851338)
(700,	-0.848281)
(750,	-0.840507)
};
\addlegendentry{Bayesian};

\addplot+[solid ,black,thick, mark options={black!50}] coordinates {
(150,	-1.13526)
(200,	-1.04789)
(250,	-1.01318)
(300,	-0.978555)
(350,	-0.968083)
(400,	-0.950671)
(450,	-0.92945)
(500,	-0.928194)
(550,	-0.906013)
(600,	-0.891075)
(650,	-0.877307)
(700,	-0.862388)
(750,	-0.850656)
};
\addlegendentry{MDL};
\end{axis}
\end{tikzpicture}}
	\end{subfigure}%
	\begin{subfigure}{0.45\textwidth}
		\centering
		\scalebox{0.6}{\begin{tikzpicture}
\begin{axis}[
		xlabel = Number of Samples,
		ylabel = Number of Edges,
		legend style={at={(0.99,0.02)},anchor=south east},
		ymin= 2, ymax= 7
]
\addplot+[solid ,black,thick, mark options={black}] coordinates {
(150, 3.34)
(200, 3.48)
(250, 3.86)
(300, 4.36)
(350, 4.6)
(400, 5.12)
(450, 5.6)
(500, 5.8)
(550, 5.98)
(600, 5.98)
(650, 6.0)
(700, 6.0)
};
\addlegendentry{Bayesian};

\addplot+[solid ,black,thick, mark options={black!50}] coordinates {
(150, 2.3)
(200, 2.88)
(250, 3.12)
(300, 3.42)
(350, 3.64)
(400, 3.84)
(450, 4.4)
(500, 4.5)
(550, 5.06)
(600, 5.48)
(650, 5.82)
(700, 5.98)
(750, 6.0)
};
\addlegendentry{MDL};
\end{axis}
\end{tikzpicture}}
	\end{subfigure}

	\caption{The average test log-likelihood and the average number of discretization edges for $\cX$ as the number of training samples is varied.
  The complete dataset size is fixed at \num{800} samples, drawn randomly from the synthetic distribution.
  The size of the subset used for training is varied, and the remaining samples are used as the test set.
  }
	\label{fig:synthetic}
\end{figure}
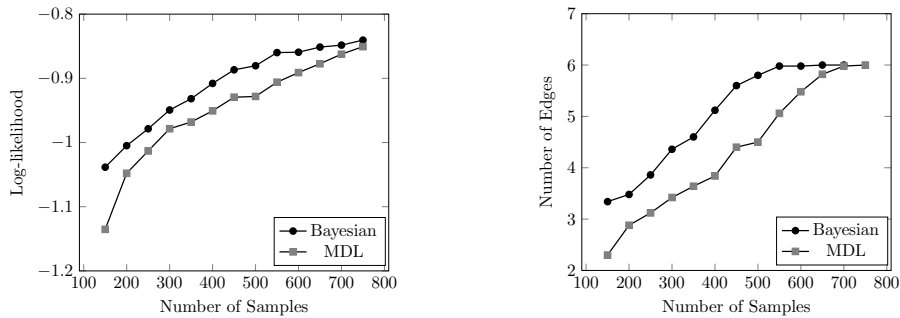

Figure~\ref{fig:synthetic} shows how the test likelihood and the number of discretization edges for $\cX$ vary with respect to the number of training samples.
A dataset of \num{800} samples is drawn from the synthetic distribution.
The size of the subset used for training is varied, and the remaining samples are used as the test set.
When the training set is small, it is difficult to recognize the underlying pattern and discretize the continuous variable $\cX$ accordingly.
In this case, both the likelihood of the training data and number of discretization edges are small.
However, the Bayesian method is more sensitive to variation in the data and produces a distribution both with higher likelihood and more discretization edges.
As the size of the training set grows, both methods converge towards the optimal discretization policy.
The Bayesian method finds the six discretization edges faster than the MDL method and consistently outperforms in likelihood.
The curves in Figure~\ref{fig:synthetic} demonstrate that the Bayesian method is indeed more sensitive to the dataset than the MDL method.
The results in Figure~\ref{fig:synthetic} are averaged over one hundred trials.



\section{Conclusion}
This paper introduced a principled discretization method for continuous variables in Bayesian networks with quadratic complexity instead of the cubic complexity of other standard techniques.
Empirical demonstrations show that the proposed method is superior to the state of the art.
In addition, this paper shows how to incorporate existing methods into the structure learning process to discretize all continuous variables and simultaneously learn Bayesian network structures.
The proposed method was incorporated and its superior performance was empirically demonstrated.
Future work will investigate edge positions at locations other than the midpoints between samples and will extend the approach to cluster categorical variables with very many levels.
All software is publicly available at \url{github.com/sisl/LearnDiscreteBayesNets.jl}.





\section*{Appendix A.}
\label{sec:appendixA}

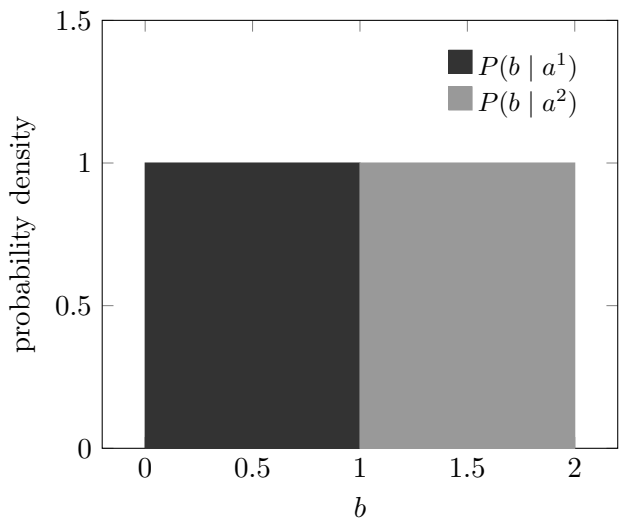
\begin{figure}[ht]
    \centering
    \vspace{1em}
    	\begin{tikzpicture}
    	\begin{axis}[xlabel=$b$, ylabel=probability density, ymin=0, ymax=1.5,
    	legend style={
    		draw=none,
    		at={(0.95,0.85)},
    		anchor=east,
    		legend columns=1,
    		font=\small,
    		/tikz/every even column/.append style={column sep=0.5cm},
    		/pgfplots/legend image code/.code={%
    			\draw[##1,/tikz/.cd,yshift=-0.25em]
    			(0cm,0cm) rectangle (0.8em,0.8em);
    		},
    	},
    	]
    	\addplot+[ybar interval, mark=none, black!80, fill=black!80] plot coordinates
    	{(0,1) (1,1)};
    	\addplot+[ybar interval, mark=none, black!40, fill=black!40] plot coordinates
    	{(1,1) (2,1)};
    	\legend{$P(b\mid a^1)$, $P(b\mid a^2)$}
    	\end{axis}
    	\end{tikzpicture}
    \caption{The resulting conditional probability distribution for $B$. Marginal methods see one uniform distribution.}
    \label{fig:marginal_counter}
\end{figure}

This section illustrates the advantage of structure-aware discretization methods over marginal discretization methods \citep{scott1979optimal,freedman1981histogram,knuth2013optimal,scargle2013studies}.
Marginal discretization methods ignore intervariable dependencies and can thus lead to suboptimal discretizations for the joint distribution.
Consider a synthetic dataset with two variables, discrete $A$ and continuous $B$, with \num{500} data samples sampled from the Bayesian network $A \rightarrow B$ described in Table~8.  
Following the experiment procedure in Section~\ref{sec:experiments}, both the MDL method and the proposed Bayesian approach find the true discretization edge for $B$ near $1.0$ and have log likelihood \num{-0.023}.
They recover the discretization shown in Figure~\ref{fig:marginal_counter}.
The Bayesian Blocks method~\citep{scargle2013studies}, as with any marginal method, cannot distinguish between the different conditional distributions and thus does not locate a discretization edge.
The resulting log likelihood is worse, at \num{-0.69}.
This example shows that structure-aware discretization methods can leverage additional information and thus produce better joint representations.

\begin{table}
	\centering
	\begin{tabular}{cl}
		\toprule
		Variable               & Distribution                                   \\ \midrule
		\multirow{2}{*}{$A$}&$P(a^1) = 0.5$     \\
		& $P( a^2) = 0.5$     \\ \midrule
		
		\multirow{2}{*}{$B$} & $p(b \mid A = a^1) \sim U[0,1]$\\
		& $p(b \mid A = a^2) \sim U[1,2]$\\
		\bottomrule
	\end{tabular}
	\vspace{1cm}
	\caption{Probability distributions for $A \rightarrow B$, in which $A$ is discrete and $B$ is continuous. The marginal distribution for $B$ is $U[0,2]$.}
\label{table:marginal_counter}
\end{table}

\section*{Appendix B.}
\label{sec:appendixB}

This section gives two examples of Bayesian networks where MDL produces comparable results to the Bayesian method.
The first example used the Auto MPG dataset.
Instead of testing on the network obtained using initially uniform discretization and structure learning, the tested network is obtained by Algorithm \ref{alg:structure_learn}, which can be considered better-structured than the network in Figure \ref{fig:auto_graph_1}.
The second example is on the Iris dataset \citep{Lichman_2013}.
In the work of \citet{Friedman_1996}, the Iris dataset was tested on the naive Bayes network structure and the prediction accuracy was shown.
The same test is reproduced below to compare the Bayesian method to the MDL discretization method.

\subsection*{B.1 Auto MPG with Comparable MDL Performance}
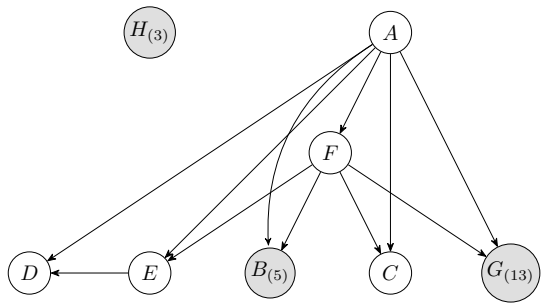
\begin{figure}[ht]
  \centering
  \scalebox{0.8}{
%
%
%
%


\begin{tikzpicture}[
      >={Stealth[round]}
      ]
      
      \node[obs]     at  (0,0)           (c2){$B_{{(5)}}$};
      \node[latent]  at  (2,0)           (c3){$C$};
      \node[latent]  at  (-2,0)         (c5){$E$};
      \node[latent]  at  (-4,0)         (c4){$D$};
      \node[latent]  at  (1,2)     (c6){$F$};
      \node[latent]  at  (2,4)     (c1){$A$};
      \node[obs]     at   (4,0)          (c7){$G_{(13)}$};
      \node[obs]     at   (-2,4)        (c8){$H_{(3)}$};

  \edge {c1}{c3,c7,c6,c5,c4};
  \draw[->,>=stealth', shorten >= 1pt] (c1) to[bend right](c2);
  \edge {c5}{c4};
  \edge {c6}{c5,c2,c3,c7};

\end{tikzpicture}

}
  \caption{A Bayesian network structure for the Auto MPG dataset that leads to comparable discretization policies for the MDL and Bayesian methods.}
  \label{fig:auto_graph_3}
\end{figure}

The Bayesian network structure in Figure~\ref{fig:auto_graph_3} and the discretization results in Table~\ref{table:auto_disc_table_3} demonstrate that the MDL method does not always produce fewer discretization intervals than the Bayesian approach and that the MDL method can produce discretization policies with comparable likelihood.
The Bayesian network in Figure~\ref{fig:auto_graph_3} was obtained by running Algorithm~\ref{alg:structure_learn} fifty times on the Auto MPG dataset with a maximum of two parents per variable.
The network is structured more favorably with regards to discretization than the network in Figure~\ref{fig:auto_graph_1}, as the discretization policies were simultaneously learned with the network structure, resulting in higher data likelihood.
Many edges appear in both discretization results and are bolded in Table~\ref{table:auto_disc_table_3}.

\begin{table}[ht]
  \centering
  \caption{
    The discretization policies for the Auto MPG dataset with the fixed structure shown in Figure~\ref{fig:auto_graph_3}.
    The methods have the same discretization for variable $A$, and the log-likelihood for the Bayesian method is only slightly better than that for the MDL method.
  }
\scriptsize
\begin{tabular}{@{}ccc@{}}
    \toprule
    Variable & Bayesian & MDL \\
    \midrule
    $A$ & \textbf{17.65}, \textbf{22.75} & \textbf{17.65}, \textbf{22.75} \\
    $C$ & \num{70.5}, \textbf{159.5}, \textbf{259.0}, \num{284.5} & \textbf{159.5}, \textbf{259.0} \\
    $D$ & \textbf{99.0}, \num{127.0} & \num{84.5}, \textbf{99.0}, \num{123.5} \\
    $E$ & \num{2764.5}, \num{3030.0}, \num{3657.5} & - \\
    $F$ & \num{14.05} & - \\
    \midrule
    Log-Likelihood & \num{-25.94} & \num{-26.83} \\
    \bottomrule
\end{tabular}
  \label{table:auto_disc_table_3}
\end{table}

\subsection*{B.2 Reproducing MDL Results from the Literature}

The Iris dataset contains variables related to the morphologic variation of three Iris flower species.
The dataset has \num{150} samples over five variables.
Only variables $E$ is discrete-valued, and corresponds to the species category.
The following experiment was conducted on the naive Bayes structure: variable $E$ is the parent of variable $A$, $B$, $C$, and $D$. Table \ref{table:iris_table_1} shows the discretization edges, the prediction accuracy of the learned classifiers, and the mean log-likelihood.
The discretization policy shown is for running the algorithms on the full dataset; the likelihood and prediction accuracies were found with 10-fold cross validation.
The MDL discretization method and the Bayesian method have the same discretization result: all discretization edges coincide. In addition, both methods have $94 \%$ accuracy, which matches the number in the literature \citep{Friedman_1996}.
The Bayesian method likelihood is slightly higher across folds than the MDL method likelihood.

\begin{table}[ht]
  \centering
  \caption{
    The discretization policies for the Iris dataset with the naive Bayes structure.
    The two methods produce the same discretization policy on each continuous variable and have nearly identical 10-fold prediction accuracy and log-likelihood.
  }
\scriptsize
\begin{tabular}{@{}ccc@{}}
    \toprule
    Variable & Bayesian & MDL \\
    \midrule
    $A$ & \num{5.45}, \num{6.15} & \num{5.45}, \num{6.15} \\
    $B$ & \num{2.95}, \num{3.35} & \num{2.95}, \num{3.35} \\
    $C$ & \num{2.45}, \num{4.75} & \num{2.45}, \num{4.75} \\
    $D$ & \num{0.8}, \num{1.75} & \num{0.8}, \num{1.75} \\
    \midrule
    Prediction Accuracy & \num{0.94} & \num{0.94} \\
    \midrule
    Log-Likelihood & \num{-2.25} & \num{-2.50} \\
    \bottomrule
\end{tabular}
  \label{table:iris_table_1}
\end{table}

\vskip 0.2in
\bibliography{my_bib}
\bibliographystyle{theapa}

\end{document}